\documentclass[sigplan,screen]{acmart}

\copyrightyear{2026}
\acmYear{2026}
\setcopyright{cc}
\setcctype{by-nc-nd}
\acmConference[ASPLOS '26]{Proceedings of the 31st ACM International Conference on Architectural Support for Programming Languages and Operating Systems, Volume 2}{March 22--26, 2026}{Pittsburgh, PA, USA}
\acmBooktitle{Proceedings of the 31st ACM International Conference on Architectural Support for Programming Languages and Operating Systems, Volume 2 (ASPLOS '26), March 22--26, 2026, Pittsburgh, PA, USA}
\acmPrice{}
\acmDOI{10.1145/3779212.3790161}
\acmISBN{979-8-4007-2359-9/2026/03}

\usepackage{xcolor}
\usepackage{amsmath}
\usepackage{amsfonts}
\usepackage{pifont}
\usepackage{multicol}
\usepackage{multirow}
\usepackage{subcaption}
\usepackage{array}
\usepackage{hyperref}
\usepackage{enumitem}
\usepackage{algorithm}
\usepackage{algpseudocode}
\usepackage{indentfirst}

\newcolumntype{C}[1]{>{\centering\arraybackslash}p{#1}} 
\usepackage{makecell}

\newcolumntype{L}[1]{>{\raggedright\let\newline\\\arraybackslash\hspace{0pt}}m{#1}}
\newcolumntype{R}[1]{>{\raggedleft\let\newline\\\arraybackslash\hspace{0pt}}m{#1}}

\newcommand{\secref}[1]{Sec.~\ref{#1}}
\newcommand{\figref}[1]{Fig.~\ref{#1}}

\AtBeginDocument{%
  }

\newif\ifhongxiang
\hongxiangfalse

\newcommand{\fasttts}{\textit{FastTTS}}

\usepackage{tikz}
\usepackage{fancyhdr}

\fancypagestyle{firstpagebadges}
{
    \fancyhead{} 
    
    \begin{tikzpicture}[remember picture,overlay]
        \node [xshift=165mm,yshift=-10mm] at (current page.north west) 
            {\href{https://www.acm.org/publications/policies/artifact-review-and-badging-current}{\includegraphics[width=1.6cm]{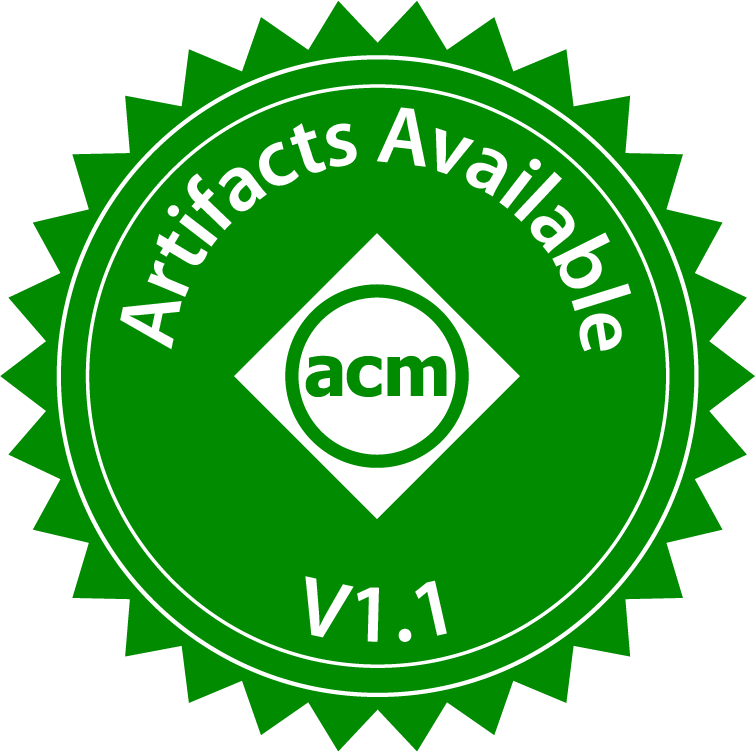}}};
        \node [xshift=183mm,yshift=-10mm] at (current page.north west) 
            {\href{https://www.acm.org/publications/policies/artifact-review-and-badging-current}{\includegraphics[width=1.6cm]{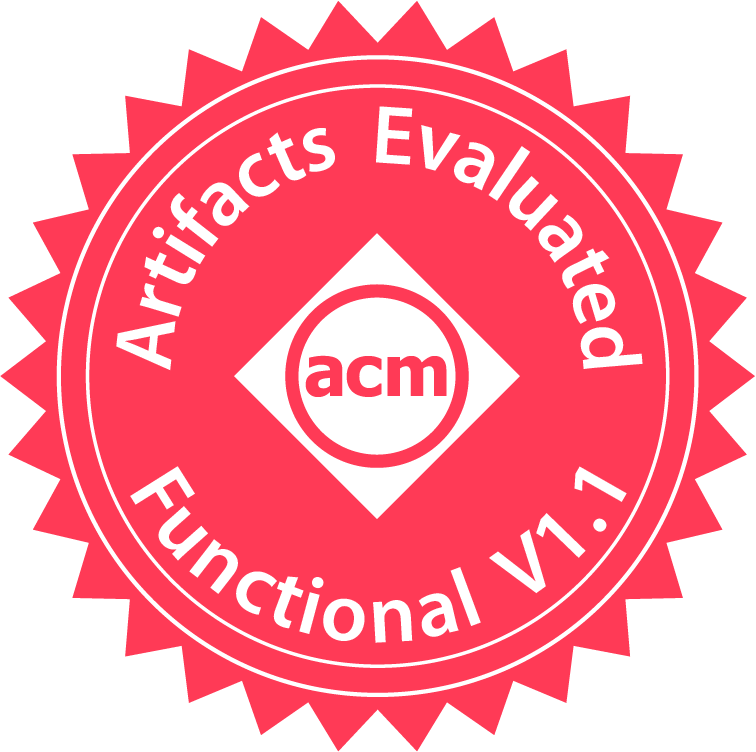}}};
        \node [xshift=201mm,yshift=-10mm] at (current page.north west) 
            {\href{https://www.acm.org/publications/policies/artifact-review-and-badging-current}{\includegraphics[width=1.6cm]{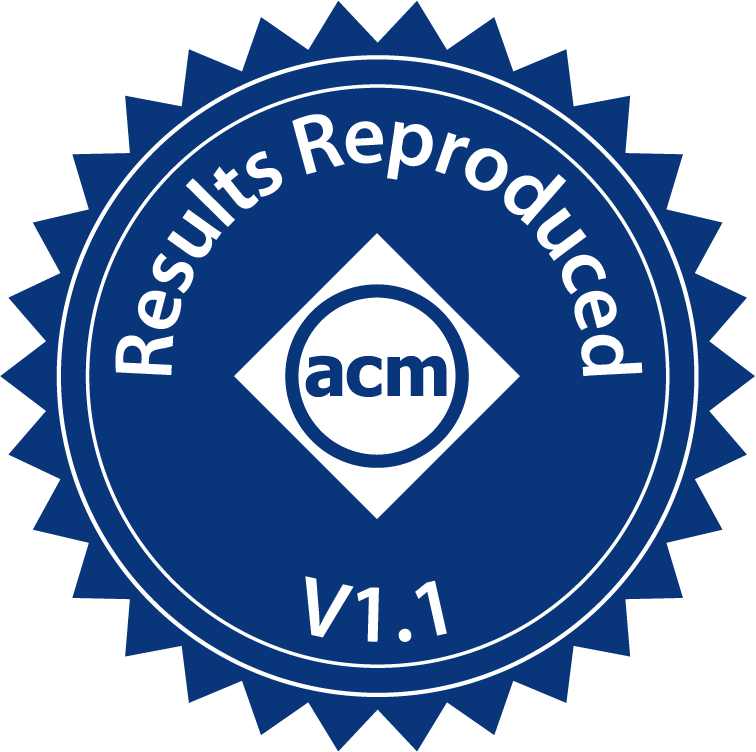}}};
    \end{tikzpicture}
}

\begin{document}

\title[FastTTS]{FastTTS: Accelerating Test-Time Scaling for \\ Edge LLM Reasoning}

\author{Hao Mark Chen}
\affiliation{%
  \institution{Imperial College London}
  \city{London}
  \country{UK}
}
\email{hao.chen20@imperial.ac.uk}

\author{Zhiwen Mo}
\affiliation{%
  \institution{Imperial College London}
  \city{London}
  \country{UK}
}
\email{zhiwen.mo25@imperial.ac.uk}

\author{Guanxi Lu}
\affiliation{%
  \institution{Imperial College London}
  \city{London}
  \country{UK}
}
\email{guanxi.lu22@imperial.ac.uk}

\author{Shuang Liang}
\affiliation{%
  \institution{Imperial College London}
  \city{London}
  \country{UK}
}
\email{shuang.liang@imperial.ac.uk}

\author{Lingxiao Ma}
\affiliation{%
  \institution{Microsoft Research}
  \city{Beijing}
  \country{China}
}
\email{lingxiao.ma@microsoft.com}

\author{Wayne Luk}
\affiliation{%
  \institution{Imperial College London}
  \city{London}
  \country{UK}
}
\email{w.luk@imperial.ac.uk}

\author{Hongxiang Fan}
\affiliation{%
  \institution{Imperial College London}
  \city{London}
  \country{UK}
}
\email{hongxiang.fan@imperial.ac.uk}

\begin{abstract}
Recent advances in reasoning Large Language Models (LLMs) are driving the emergence of agentic AI systems.
Edge deployment of LLM agents near end users is increasingly necessary to protect data privacy, enable offline use, and provide responsive interaction with local context.
However, strict memory constraints on edge devices limit deployment to smaller LLMs, whose reasoning capabilities are much weaker than those of large cloud models, hindering practical deployment of edge agentic AI.
Test-Time Scaling (TTS) offers a promising solution by allocating more compute during inference to enhance the reasoning capability of edge LLMs. However, current TTS methods introduce heavy hardware performance overhead on resource-constrained devices, making them impractical for real applications. 
To address this challenge, we present \textit{FastTTS}, a serving system that enables fast and efficient TTS for memory-constrained LLM reasoning.
After analyzing common patterns across various TTS methods and identifying their performance bottlenecks,
we introduce three novel techniques: \textit{i)} Speculative Beam Extension, which mitigates system stragglers caused by irregular reasoning paths, \textit{ii)} Asymmetric Multi-Model Memory Allocation, which dynamically balances memory usage between token generation and reasoning-step verification, and \textit{iii)} Dynamic Prefix-Aware Scheduling, which optimizes reasoning execution to maximize KV-cache reuse across search paths.
\textit{FastTTS} offers a plug-and-play third-party library on top of vLLM, enabling edge LLMs ($\leq$ 7B) on a single consumer GPU (24 GB) to match cloud-model accuracy and cloud-measured latency.
Comprehensive evaluation shows that \textit{FastTTS} achieves an average 2.2$\times$ higher goodput and reduces latency by 38\%--68\% compared to the vLLM baseline; it pushes the boundaries of low-latency Test-Time Scaling on memory-constrained edge devices and highlights the potential for democratizing agentic AI.
\end{abstract}




\renewcommand{\shortauthors}{Hao Mark Chen et al.}

\maketitle
\thispagestyle{firstpagebadges}

\section{Introduction}

\begin{figure}
  \centering
  \begin{subfigure}[t]{0.41\linewidth}
    \includegraphics[height=0.16\textheight, width=\linewidth]{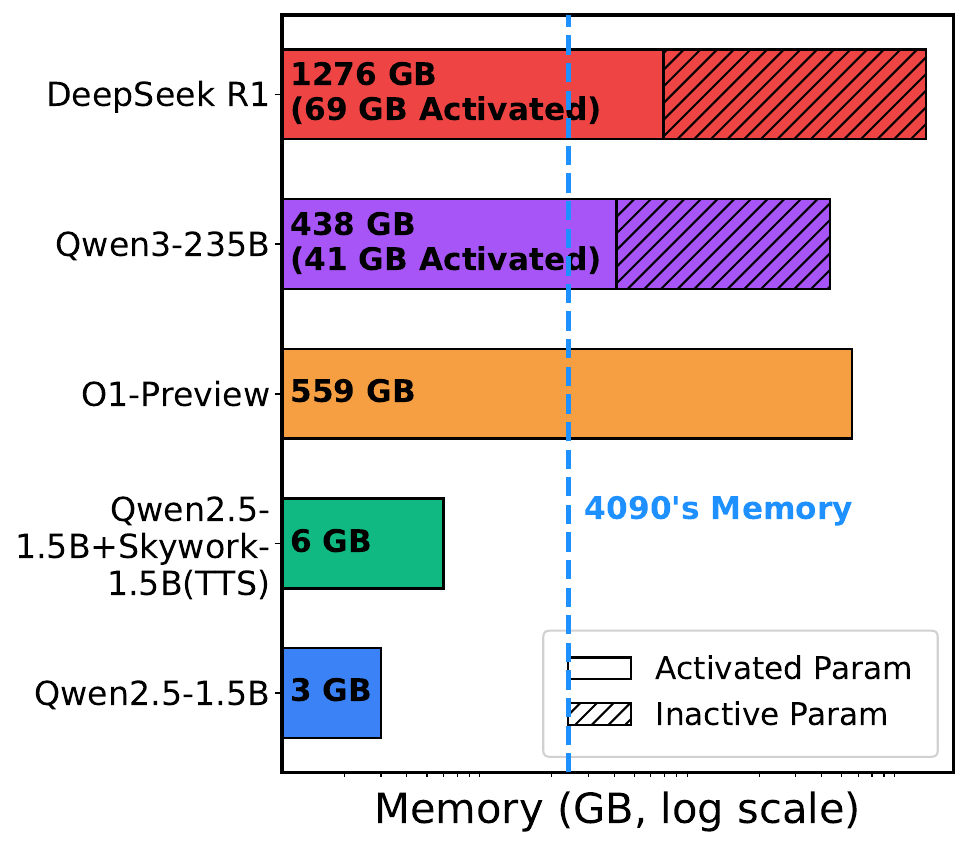}
    \caption{Memory constraints}
    \label{fig:scaling-left}
  \end{subfigure}
  \begin{subfigure}[t]{0.58\linewidth}
    \includegraphics[height=0.16\textheight, width=0.93\linewidth]{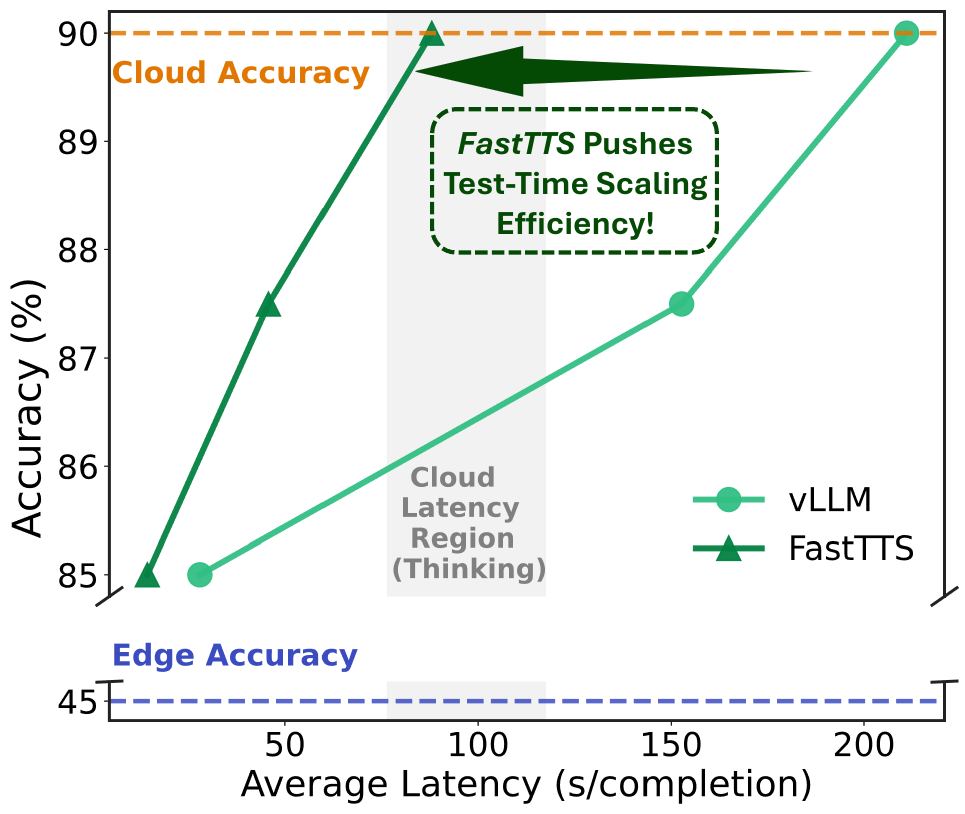}
    \caption{Reduced latency by \fasttts{}}
    \label{fig:scaling-right}
  \end{subfigure}\hfill
  \caption{
  (a) Memory cost across models. (b) \fasttts{} enables low-latency edge deployment of reasoning models. Cloud accuracy: GPT-o1-preview. Edge accuracy: Qwen2.5-Math-1.5B. Cloud latency from the first-answer latency of GPT-o3-pro and GPT-5 (thinking models)~\cite{artificialanalysis_leaderboard_reasoning_large, yang2025reasonflux}.
  }
  \label{fig:scaling-pair}
  \Description{}
\end{figure}

Recent advances in reasoning LLMs have unlocked significant progress in solving complex tasks such as multi-hop question answering, tool use, and long-horizon planning~\cite{deepmind2025alphaevolve, guo2025deepseekr1, yang2025qwen3}. These capabilities are foundational for agentic AI systems, where AI agents can plan, act, and interact autonomously. As such systems move closer to real-world deployment, there is a growing demand to deploy strong reasoning LLMs at the edge (e.g., on AI PCs), where agentic systems can preserve data privacy, enable personalization, operate offline or with limited connectivity, and interact with local environments using high-level intelligence.
However, edge hardware imposes severe memory limitations (e.g., a single consumer GPU with 8–24 GB VRAM), restricting deployment to edge LLMs ($\leq$ 7B) that cannot match the reasoning performance of large cloud models, limiting their effectiveness in complex tasks.
As shown in~\figref{fig:scaling-left}, the memory capacity of consumer-grade GPUs restricts deployment to models like Qwen2.5-Math-1.5B, resulting in a significant gap in reasoning ability compared to large-scale cloud LLMs.

Deploying strong reasoning LLMs on the edge is essential for realizing \textbf{democratized agentic AI}, where intelligent agents are decentralized and run directly on client-side devices for better privacy and local integration.
Such guarantees are particularly critical in sensitive domains such as healthcare, autonomous driving, and defense, where on-device reasoning is essential for privacy, energy efficiency, or responsiveness~\cite{nissen2025medicine, saad2025intelligence}.
To achieve this,
Test-Time Scaling (TTS)~\cite{snell2024scaling, beeching2024scalingtesttimecompute} has recently emerged as a promising candidate to bridge the reasoning gap between small, edge-deployable LLMs and large cloud-based models. 
Instead of relying on scaling model parameters during training, TTS allocates additional compute during inference to improve generation and reasoning quality.
Despite its great potential, deploying TTS naively on existing serving systems incurs significant latency overhead, making it impractical for real-time applications. 
As shown in ~\figref{fig:scaling-right}, using a baseline vLLM implementation to match the accuracy of a large cloud model results in 200 seconds of latency, nearly doubling the latency of large models on cloud infrastructure. 
To realize the vision of democratized agentic AI, there is an urgent need for an efficient, edge-ready serving infrastructure that makes TTS both performant and practical.

To build such a system, this work begins by first analyzing mainstream TTS methods and abstracting their common execution patterns~(\secref{motivation:tts}). 
We observe that most TTS methods follow a common verifier-guided search pattern that iteratively expands a tree of reasoning paths, where different TTS methods can be viewed as variants or subsets of this approach.
Building on this finding, we next conduct a systematic profiling of this common pattern in TTS methods to identify the system-level bottlenecks that hinder its efficiency.
Our analysis reveals the following three challenges:
\begin{itemize}[leftmargin=*]
\item \textbf{\textit{Challenge-1: Hardware Underutilization from Irregular Search Paths.}}
Advanced TTS methods employ multi-step, verifier-guided generation, where each search path may produce a variable number of tokens per reasoning step. This divergence leads to execution stragglers, causing idle GPU resources and severely degrading hardware utilization. (\secref{motivation:spec_exe})
\item \textbf{\textit{Challenge-2: Suboptimal Exploitation of Dynamic Prefix Sharing.}} The parallel search in TTS creates substantial opportunities for prefix-caching reuse, as many generation paths share common thinking prefixes. 
However, these sharing patterns are dynamic and only known at run-time. 
Naive scheduling ignores this locality, causing KV cache eviction and re-computation, which is especially severe on memory-constrained edge devices. (\secref{motivation:prefix})
\item \textbf{\textit{Challenge-3: Constrained Memory for Multi-Model Execution.}} A core component of many TTS methods is the use of a separate verifier model to guide the generator. This requires collocating two distinct models in the constrained memory of a consumer-grade GPU. 
It leads to higher latency due to limited batch size, thereby undermining the benefits of TTS. (\secref{motivation:memory})
\end{itemize}

To overcome these obstacles, we present \textbf{\fasttts{}}, a serving system that integrates three synergistic optimizations to make TTS practical on edge devices. To address \textit{Challenge-1}, we introduce \textbf{Speculative Beam Extension} that generates speculatively to hide the latency of irregular workloads. To tackle \textit{Challenge-2} and \textit{Challenge-3}, \fasttts{} combines two memory-aware optimizations: \textbf{Dynamic Prefix-Aware Scheduling} reorders execution to maximize KV cache reuse from dynamic prefix sharing, and \textbf{Asymmetric Multi-Model Memory Allocation} intelligently partitions memory between the generator and verifier to improve throughput. Together, we push the boundaries of edge deployment of TTS (\figref{fig:scaling-right}), making fast and high-quality reasoning feasible on memory-constrained edge devices.

The main contributions of this paper are threefold:
\begin{itemize}[leftmargin=*]
    \item We systematically analyze the common execution patterns of modern verifier-guided TTS methods and identify their core system-level bottlenecks with a comprehensive performance profiling.
    \item We design and implement \textbf{\fasttts{}}, a high-performance serving system for TTS that incorporates three novel and synergistic optimizations: Speculative Beam Extension, Dynamic Prefix-Aware Scheduling, and Asymmetric Multi-Model Memory Allocation.
    \item We conduct a comprehensive evaluation on representative edge hardware, demonstrating that \fasttts{} achieves an average 2.2$\times$ higher goodput and reduces the latency by 38\%--68\% compared to the vLLM baseline.
\end{itemize}

\section{Background}\label{sec:motivation}

\subsection{LLM Reasoning}

Reasoning is a critical capability for Large Language Models (LLMs), enabling multi-step problem solving and complex decision-making. This reasoning capability is initially established through reinforcement learning methods such as Guided Reinforcement Policy Optimization (GRPO), as exemplified by DeepSeek-R1~\cite{guo2025deepseekr1}. Such RL training fosters emergent abilities like long Chain-of-Thought (CoT) reasoning, which in turn expands the applicability of LLMs to domains including mathematical problem solving~\cite{shao2024deepseekmath}, scientific discovery~\cite{deepmind2025alphaevolve, mirza2025framework, yang2024qwen2}, coding assistant~\cite{guo2024deepseek, gai2025exploring}, and multi-hop question answering~\cite{yang2025qwen3, guo2025deepseekr1}.

\subsection{Test-Time Scaling (TTS) Methods}

\begin{figure}[t]
    \centering    \includegraphics[width=0.99\linewidth]{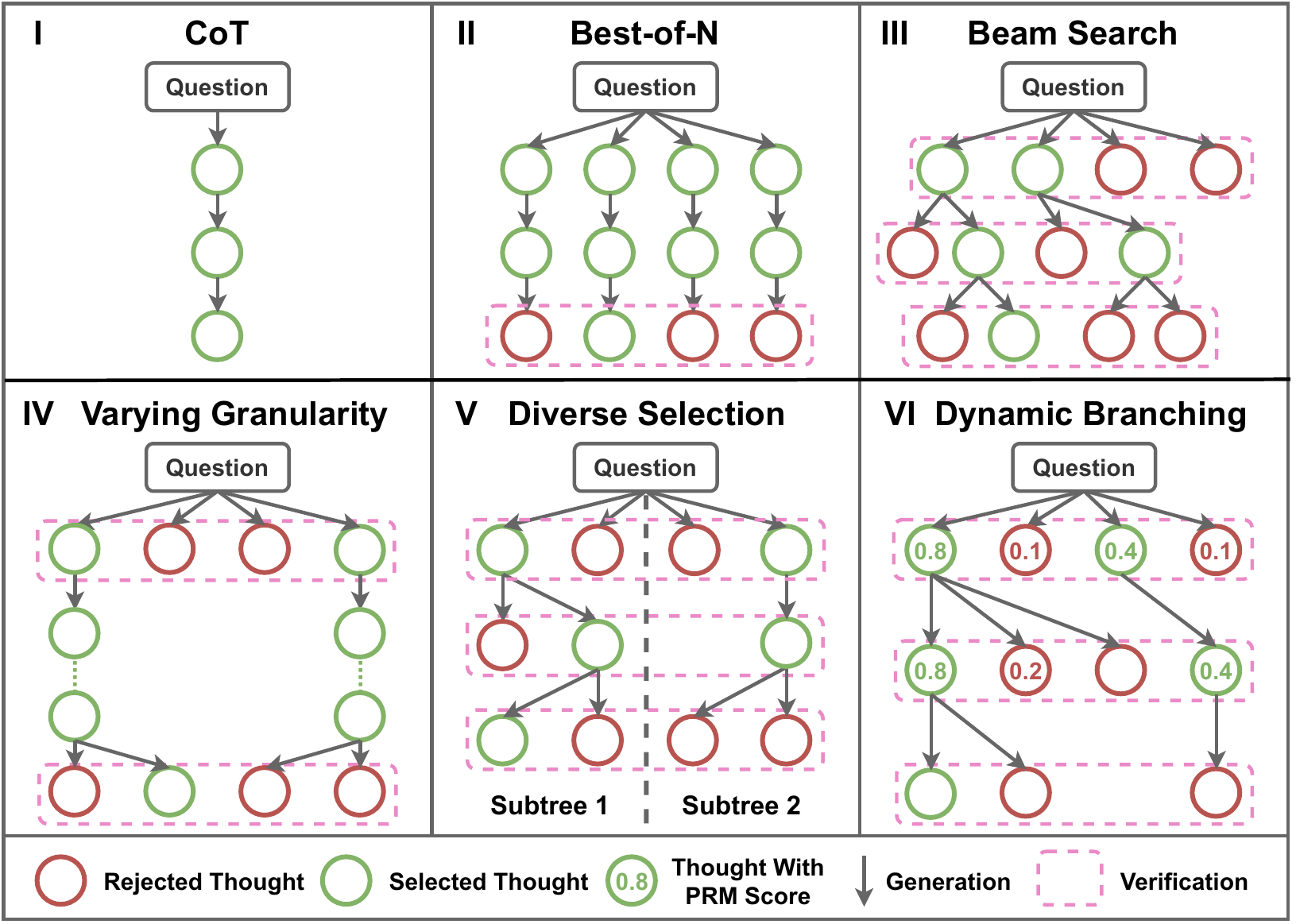}
    \caption{
    Illustration of different TTS methods.
    }
    \label{fig:tts_illustration}
  \Description{}
\end{figure}

While long Chain-of-Thought (CoT) reasoning enhances the capabilities of LLMs, smaller models still lag significantly behind their larger counterparts~\cite{liu2025can}. To bridge this gap, TTS increases the computational budget for inference by exploring multiple reasoning paths in parallel~\cite{snell2024scaling, beeching2024scalingtesttimecompute, liu2025can}. Early TTS methods mainly relied on Best-of-N (BoN) sampling, where an Outcome Reward Model (ORM) selects the best solution from a set of fully generated candidates~\cite{wang2022self, cobbe2021training}. However, BoN offers limited guidance during generation and yields less diversity in reasoning path structures (~\figref{fig:tts_illustration}).
The introduction of Process Reward Models (PRMs), which evaluate intermediate reasoning steps, has enabled advanced verifier-guided search algorithms such as Beam Search and Monte Carlo Tree Search (MCTS)~\cite{lightman2023let, uesato2022solving, snell2024scaling}. These methods follow a generation--verification paradigm: a PRM periodically scores partial solutions, expanding high-scoring trajectories and pruning weak ones, thereby concentrating computation on promising paths~\cite{hooper2025ets, wu2024inference, chen2025rethinking}. As a result, the LLM produces a diverse reasoning \emph{tree} rather than a single chain.

PRMs are primarily categorized as either discriminative or generative~\cite{RewardBench2}. A discriminative PRM functions as a sequence classifier; in a single forward pass, it takes a full reasoning path as input and outputs a score for each intermediate step~\cite{skyworkopeno12024, wang2023math, lightman2023let}. In contrast, a generative PRM is an auto-regressive model that must first generate its own textual critique before providing a final score, a significantly more expensive process~\cite{zhao2025genprm}. 
Due to their superior balance of model parameters, reasoning quality, and hardware efficiency, discriminative PRMs are the preferred choice for state-of-the-art, verifier-guided TTS systems, particularly for memory-constrained edge deployment~\cite{RewardBench2}.
Hence, our system focuses on discriminative PRMs.
In contrast, as noted by~\cite{snell2024scaling, liu2025can}, multi-step lookahead approaches~\cite{chen2024toward, wang2025towards}, such as Monte Carlo Tree Search (MCTS)~\cite{feng2023alphazero}, introduce significant sampling and latency overhead with inferior accuracy, hence we do not consider them in this work.


\subsection{LLM Serving}
Serving frameworks such as vLLM~\cite{kwon2023efficient} and SGLang~\cite{zheng2024sglang} have been developed to optimize throughput and latency in streaming query scenarios.
These systems incorporate key optimizations, including KV cache management to avoid recomputing attention states, paged attention to reduce GPU memory fragmentation, and preemptive scheduling to handle memory constraints by swapping requests.
For TTS serving, goodput will be a more useful metric rather than throughput, as not all generated tokens will be selected for the final output.
Despite its importance, no serving system to date natively supports the structured, multi-path search required for TTS in reasoning tasks.


\section{Motivation}

In \secref{motivation:tts}, we analyze common patterns in recent TTS methods. Subsequently, we conduct performance profiling and identify the key performance bottlenecks in ~\secref{motivation:challenges}.

\subsection{Patterns Analysis in TTS Methods}
\label{motivation:tts}

Recent advancements in LLM reasoning have led to a variety of TTS methods, evolving from simple parallel sampling to more sophisticated search methods~\cite{zhang2025survey, chen2025rethinking, snell2024scaling, hooper2025ets, beeching2024scalingtesttimecompute}. 
As illustrated in \figref{fig:tts_illustration}, this evolution marks a structural shift in the generation process: from exploring parallel but independent chains (e.g., CoT and Best-of-N) to constructing structured reasoning trees that allow for intermediate guidance and pruning (e.g., Beam Search, MCTS, and their variants).
While recent studies show that small models can attain strong reasoning ability via long sequential CoT~\cite{guo2025deepseekr1}, tree-structured reasoning remains essential in edge settings. First, it is orthogonal to sequential CoT and can further enhance the performance of small models. Second, parallel tree-based search has been demonstrated to be substantially more latency-efficient and token-efficient than purely sequential scaling~\cite{wang2025faster, ghosal2025does}, which is critical for maintaining responsiveness under the tight resource constraints of edge deployment.

While these methods vary in their specific heuristics, they share a common underlying execution pattern: a verifier-guided search that iteratively expands a tree of reasoning paths. This process can be generalized into a two-stage loop:
\begin{enumerate}[leftmargin=*]
    \item \textbf{Generation:} From a set of active reasoning paths (\textit{beams}), the generator extends each one by generating a new thinking step, which consists of an arbitrary number of tokens.
    \item  \textbf{Verification: } A PRM, or verifier, evaluates each newly generated step and assigns a score. Top-scoring paths are then replicated to spawn the next set of active beams, while the rest are pruned.
\end{enumerate}

This two-step process repeats until all reasoning paths reach a terminal state.
Various search algorithms shown in \figref{fig:tts_illustration} can be understood as specific implementations of this general pattern, differing in the heuristics they apply during the Generation or Verification stage. 
For instance, during Verification, standard \textbf{Beam Search} selects the top-K candidates globally with a static branching factor. 
In contrast, \textbf{Diverse Selection}~\cite{snell2024scaling, beeching2024scalingtesttimecompute} modifies this to improve diversity by choosing the top candidate from distinct subtrees, while \textbf{Dynamic Branching}~\cite{hooper2025ets, wu2024inference} makes the branching factor itself adaptive to verifier scores and system state. 
Other methods, like \textbf{VG-Search}~\cite{chen2025rethinking}, instead modify the Generation stage by altering the length of the thinking steps with varying verification granularities.

\begin{figure}[t]
    \centering
    \begin{subfigure}[c]{0.4\linewidth}
        \centering
        \includegraphics[width=\linewidth]{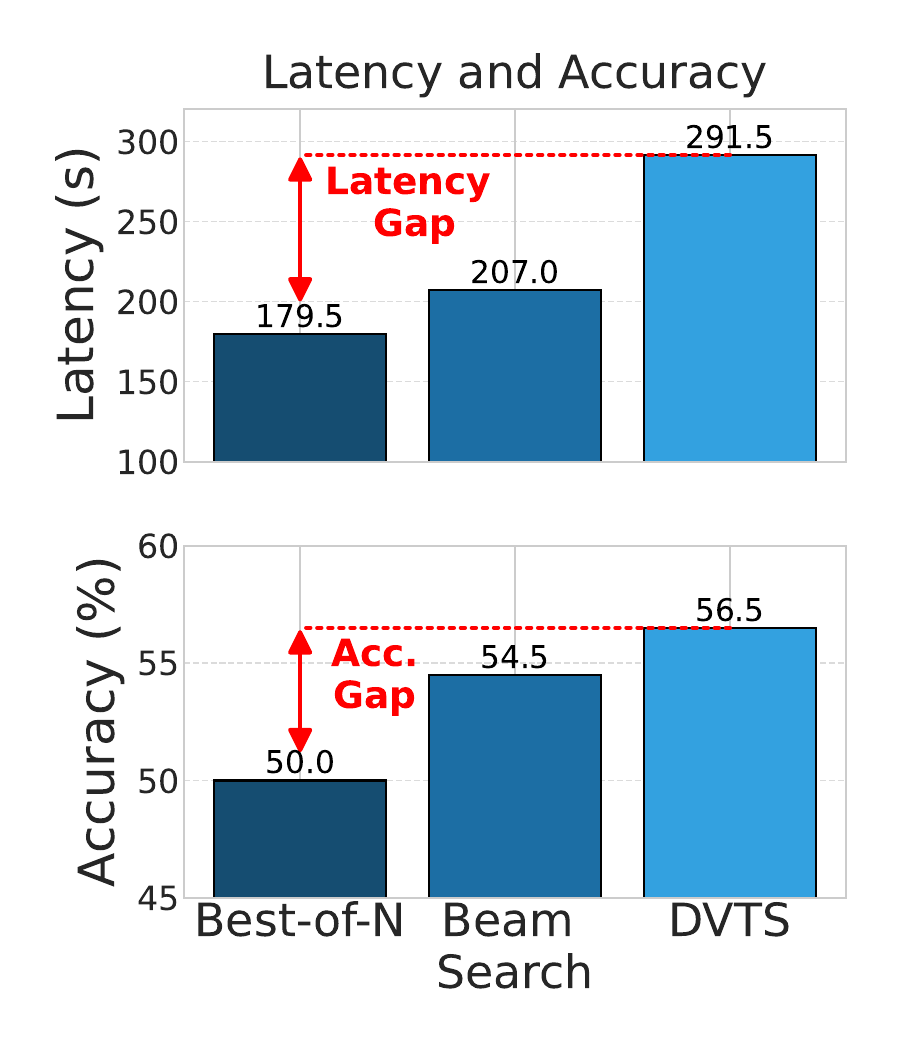}
    \end{subfigure}
    \hfill
    \begin{subfigure}[c]{0.59\linewidth}
        \centering
        \includegraphics[width=\linewidth]{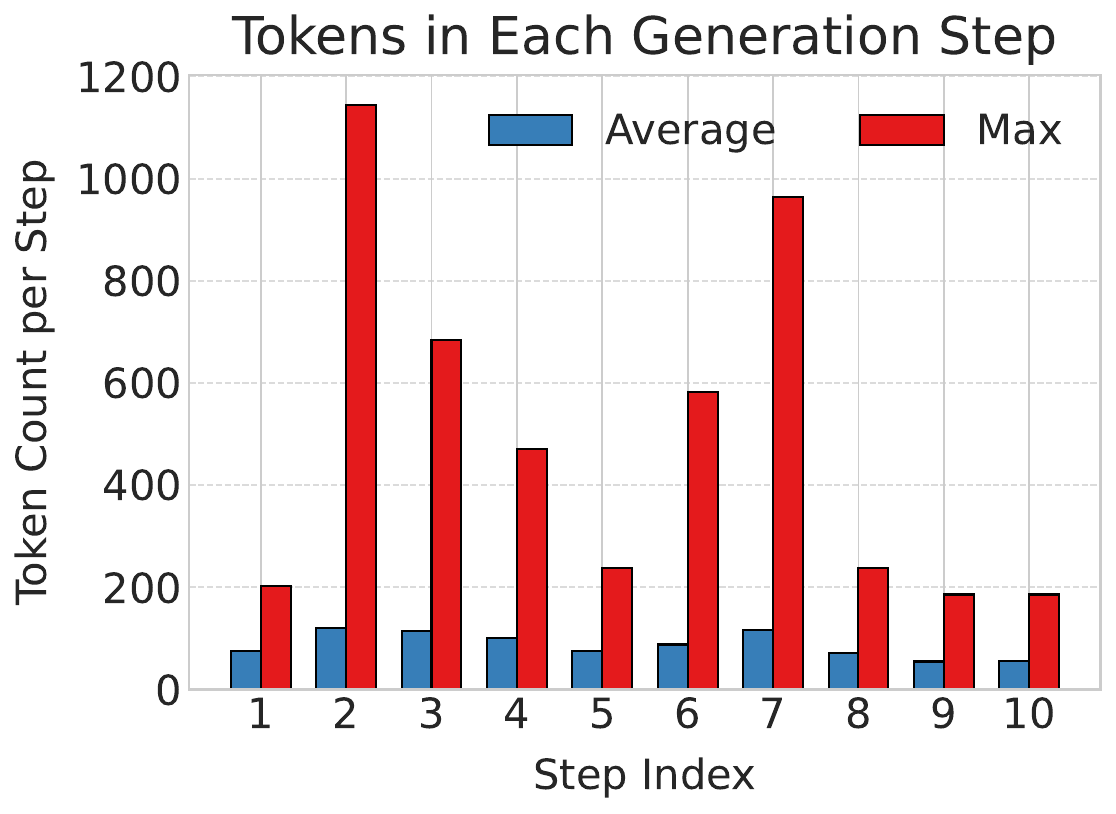}
    \end{subfigure}
    \caption{
    Left: Accuracy vs.~latency for different TTS methods on MATH-500 datasets. 
    Right: Avg. and max. token count per generation step of Qwen2.5-Math-1.5B on AIME.
    }
    
    \label{fig:tts_acc_spec_token_count}
  \Description{}
\end{figure}

To understand the accuracy--latency trade-offs across different TTS methods, we conduct evaluations on the MATH-500 dataset. 
As illustrated in \figref{fig:tts_acc_spec_token_count}~(left), while advanced search methods often achieve higher algorithm accuracy, their overall latency remains a critical bottleneck. 
To address this system-level performance gap, we analyze the challenges shared by the abstracted TTS pattern.

\subsection{Challenge Analysis and Performance Profiling}\label{motivation:challenges}
\subsubsection{Hardware Underutilization from Irregular Workloads}
\label{motivation:spec_exe}

A core challenge in serving verifier-guided TTS methods stems from the highly irregular and unpredictable workloads they create. Unlike simple token-level generation, the number of tokens generated from a thinking step between verifications can vary dramatically across parallel search paths. We analyze the distribution of these step lengths on the AIME dataset. As shown in \figref{fig:tts_acc_spec_token_count}~(right), the disparity is extreme.  
This vast difference between the average and outlier path lengths persists across all steps.

\begin{figure}[tbp]
    \centering
    \includegraphics[width=0.99\linewidth]{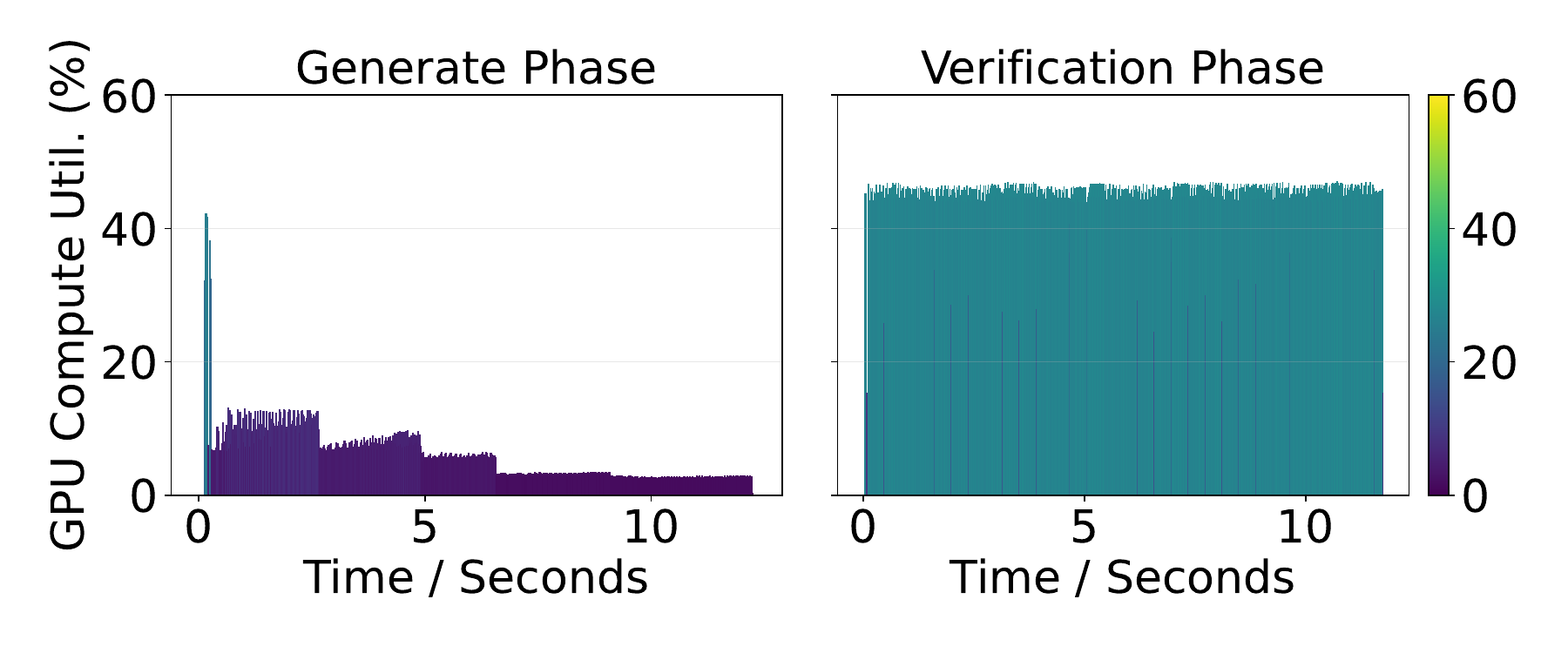}
    \caption{GPU compute utilization in generation and verification phases over time. Irregular during the generation phase. The metrics are collected using Nsight Systems, NVIDIA’s official profiling tool, at a sampling rate of 10,000 samples per second for the Tensor Core utilization metrics.
    }
    \label{fig:util_generate_encode}
\end{figure}

This workload irregularity leads directly to severe hardware underutilization. In a batch of parallel beams, the system must wait for the longest path, known as the "straggler", to complete before proceeding to the next verification stage. 
As shorter paths finish early, GPU resources are left idle, leading to inefficient resource utilization. 
\figref{fig:util_generate_encode} visualizes this problem using GPU compute utilization metrics from Nsight Systems~\cite{nvidia_nsight_systems}. During the generation phase, utilization peaks at the start but then plummets and progressively decays as more beams complete, leaving the GPU underutilized while waiting for the final straggler. This stands in stark contrast to the consistently high utilization seen during the verification phase (\figref{fig:util_generate_encode}), where workloads are uniformly prefilling. 
This issue is especially pronounced in edge settings, where small batch sizes render continuous batching inapplicable.
Such divergence leaves hardware resources idle and significantly increases end-to-end latency.


\subsubsection{Suboptimal Exploitation for Dynamic Prefix Sharing Under Limited Memory}\label{motivation:prefix}

\begin{figure}
    \centering
    \includegraphics[width=0.99\linewidth]{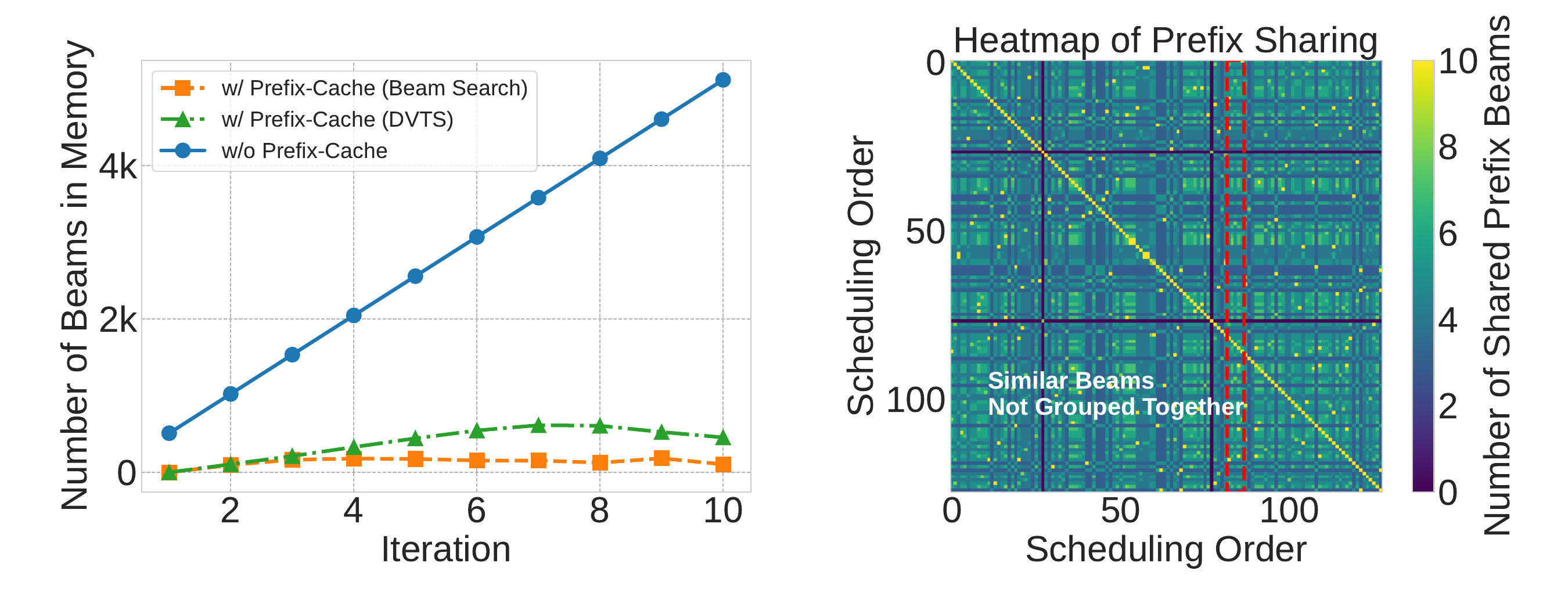}
    \caption{Optimization opportunity in Dynamic Prefix-Cache Sharing. 
    Left: Prefix-cache sharing enables potential substantial memory savings for different TTS methods.
    Right: Naive scheduling overlooks the dynamic nature of prefix-cache sharing.
    }
    \label{fig:prefix_motivation}
\end{figure}

The tree-like exploration of reasoning paths in TTS creates a significant opportunity for memory optimization through KV cache sharing, as shown in \figref{fig:prefix_motivation}~(left). 
The importance of exploiting such opportunities becomes particularly important under tight memory constraints for TTS reasoning.
Since multiple beams often share a common prefix, scheduling these beams together in a batch enables KV cache reuse and avoids frequent cache eviction. 
A scheduling policy that exploits this locality can also enable a larger effective batch size within a constrained memory budget. 
However, these prefix-sharing patterns are dynamic and only emerge at run-time as the reasoning tree expands.
The current scheduling policy does not address this, as shown in \figref{fig:prefix_motivation}~(right).
This necessitates a dedicated run-time scheduling policy that can maximize KV cache reuse, thereby minimizing redundant computation and memory access.

\subsubsection{Constrained Memory for Multi-Model Execution} \label{motivation:memory}

\begin{figure}[t]
    \centering
    \includegraphics[width=0.99\linewidth]{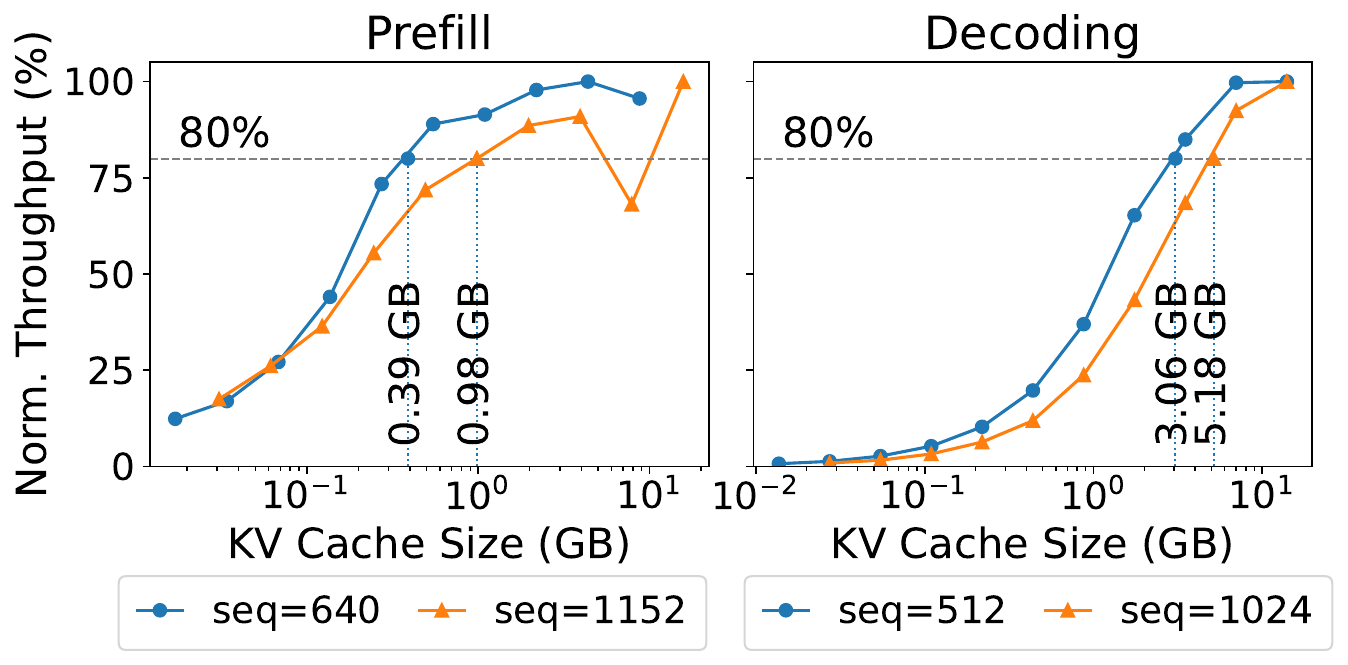}
    \caption{Normalized throughput versus KV cache size in prefill and decoding stages. Prefill saturates much easily.
    }
    \label{fig:throughput_vs_kv_cache}
\end{figure}

While TTS is deployable on edge devices with smaller models, its performance is severely hampered by constrained GPU memory on the edge (\figref{fig:scaling-right}). 
Verifier-guided search on a single device inherently requires collocating multiple models and accommodating potentially large search widths, which together place significant pressure on memory resources. 
Previous work has shown that LLM throughputs are greatly affected by available GPU memory, which determines the maximum batch size~\cite{kwon2023efficient, sheng2023flexgen}.
Addressing this bottleneck is therefore critical for improving LLM reasoning performance on edge devices.

In a memory-constrained TTS system, the generator and verifier share the same limited pool of KV cache memory. 
However, these two components exhibit vastly different throughput sensitivities to their allocated memory. 
The verifier, which processes prompts in large batches (prefill), is typically compute-bound, while the generator, which decodes tokens one by one, is memory-bandwidth bound and highly sensitive to KV cache size.
This is demonstrated in \figref{fig:throughput_vs_kv_cache}, which shows that the verifier's prefill stage reaches 80\% of its peak throughput with less than 1 GB of KV cache. In contrast, the generator's decoding stage requires 5--10$\times$ more memory to reach the same relative throughput. 
This performance asymmetry reveals a key opportunity: instead of partitioning memory arbitrarily, a carefully profiled, asymmetric allocation can significantly improve overall system throughput by providing each component with the optimal amount of memory.

\begin{figure*}[th!]
    \centering
    \includegraphics[width=0.99\linewidth]{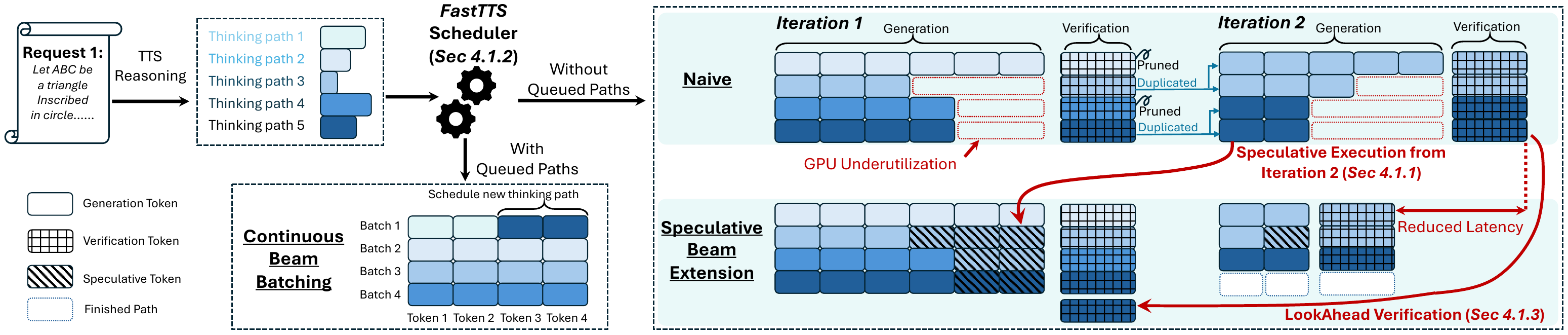}
    \caption{Speculative Beam Extension.
    }
    \label{diagram:spec}
\end{figure*}

\section{\fasttts{}: Method and Optimization}

\subsection{Speculative Beam Extension}
To mitigate the inefficiency from irregular thinking step lengths (\secref{motivation:spec_exe}), we propose \textbf{Speculative Beam Extension}, a technique that opportunistically leverages this underutilized hardware.
The key idea is to speculatively generate future tokens for beams with short thinking steps in the current iteration, effectively overlapping computations and hiding the latency of stragglers. The high-level procedure is detailed in \figref{diagram:spec} and Algorithm~\ref{alg:specbeam}.

The core logic resides in the generation \texttt{while} loop (lines 7--14), which runs until all beams ($\in B$) complete their current generation step.  
Beams selected for speculative generation are referred to as \textit{speculative candidates}.  
Within the loop, the system generates one token for both unfinished requests and speculative candidates (line 10), then updates the finished-beam set.  
From the newly finished beams, $SelectSPEC$ (line 12) chooses the most promising candidates, as detailed in \secref{sec:spec_beam_selection}.  
Once all beams are completed, the algorithm enters the standard verification and selection phase (lines 15--17). 
We verify beams without considering speculative tokens to ensure \textbf{algorithmic equivalence}.
Finally, we duplicate all selected beams for branching. If a beam underwent speculation, only its duplicates have speculative tokens truncated (lines 18–-19), while the original remains intact to simulate divergence.
The truncation length is drawn from a normal distribution with mean $R$.

\subsubsection{Speculative Candidate Selection} \label{sec:spec_beam_selection}
To maximize the benefit of speculative execution, our selection of \textit{speculative candidates} is guided by a two-fold objective: minimizing the system overhead incurred during the process, and maximizing the probability that the speculative work will be useful.


To maximize the utility of speculative execution while maintaining algorithmic equivalence, we use a low-cost heuristic to prioritize how speculative compute resources are allocated. As verifier scores between consecutive steps are often correlated~\cite{chen2025rethinking}, the score from the previous step serves as an effective, zero-overhead proxy for a beam's probability of being retained by the search algorithm. Our system policy partitions these scores into $B$ discrete bins, $\{C_1, \ldots, C_B\}$, where $C_1$ is the highest-score bin and $B$ is the search's branching factor. For a beam $b_i$ with score $s_i$, our policy determines its \textit{speculative potential}---the theoretical maximum number of branches it is eligible to generate speculatively, $M_i$:
\[
\text{If } s_i \in C_j, \quad \text{then } M_i = B - j + 1.
\]
The value $M_i$ serves as an upper bound and a scheduling priority. In practice, the actual number of speculative branches is determined opportunistically. 
To maintain a constant batch size and avoid introducing latency, speculative work is performed lazily: as standard beams in the batch complete, the newly available execution slots are filled by speculative branches from the highest-priority completed beams (i.e., those with the highest $M_i$).
The policy thus dynamically allocates a larger compute budget to the beams most likely to be chosen by the unmodified search algorithm, increasing the probability that the speculative work will be useful without altering the final outcome.

\begin{algorithm}[t]
\caption{\textit{Speculative Beam Extension}}
\label{alg:specbeam}
\begin{algorithmic}[1]
\small
\Function{SpecBeamExtend}{$B, R$}
    \State \textbf{Input:} Set of active beams $B$, Truncation Ratio $R$
    \State \textbf{Output:} Next set of beams $B_{\text{next}}$
    \State $\triangleright$ \textit{Generation with Speculation}
    \State $B_{\text{finished}} \gets \emptyset$
    \State $B_{\text{spec}} \gets \emptyset$
    \While{$B_{\text{stragglers}} \neq \emptyset$}
        \State $B_{\text{stragglers}} \gets B \setminus (B_{\text{finished}} \cup B_\text{spec})$
        \State $B_{\text{running}} \gets B_{\text{stragglers}} \cup B_{\text{spec}}$
        \State $B_{\text{new\_finished}} \gets$ \Call{GenerateOneToken}{$B_{\text{running}}$}
        \State $B_{\text{finished}} \gets B_{\text{finished}} \cup B_{\text{new\_finished}}$
        \State $B_{\text{new\_spec}} \gets$ \Call{SelectSpec}{$B_{\text{new\_finished}} \setminus B_{\text{spec}}$}
        \State $B_{\text{spec}} \gets B_{\text{spec}} \cup B_{\text{new\_spec}}$
    \EndWhile
    \State $\triangleright$ \textit{Verification and Selection}
    \State $Scores \gets$ \Call{Verifier.Evaluate}{$B$}
    \State $B_{\text{selected}} \gets$ \Call{Select}{$B, Scores$}
    \State $\triangleright$ \textit{Branching and Truncation}
    \State $B_{\text{selected}} \gets$ \Call{DuplicateThenTruncate}{$B_{\text{selected}}, R$}
    \State \Return{$B_{\text{selected}}$}
\EndFunction
\end{algorithmic}
\end{algorithm}

\subsubsection{Two-Phase Scheduling with Preemption}
To improve GPU utilization without introducing latency overhead or harming responsiveness, we introduce a two-phase, preemptible scheduling policy tailored for TTS workloads. Unlike traditional inference where continuous batching is only effective across multiple user requests, a single TTS request decomposes into many parallel reasoning paths. This unique structure allows for a special form of continuous batching \textit{within} a single request, which we term \textbf{Continuous Beam Batching}. Our scheduler leverages this opportunity in a two-phase approach:

\begin{itemize}[leftmargin=*,topsep=0pt,itemsep=2pt,parsep=0pt]
    \item \textbf{Phase 1: Continuous Beam Batching.} The scheduler's primary mode is to continuously batch the parallel thinking paths generated by the active TTS request from the waiting queue. This reduces the latency of a single request by maximizing GPU throughput of all thinking paths.

    \item \textbf{Phase 2: Speculative Execution.} When all available reasoning paths are being processed with an empty waiting queue, it transitions to the speculative phase. In this phase, it performs \textbf{Speculative Beam Extension} on completed beams to keep the execution batch full, effectively hiding straggler latency.
\end{itemize}



The speculative phase is fully preemptible: if a new request arrives or a running request is preempted due to memory constraints, all speculative generation is immediately stopped, and the system reverts to Phase~1 to serve the new request. This two-phase design ensures minimal overhead and preserves low-latency responsiveness.
Crucially, the speculative phase does not introduce extra tail latency, as all speculative executions are strictly terminated---regardless of their progress---once all the standard beam generations for the current step finish.

\subsubsection{LookAhead Verification}
A key optimization opportunity arises when Speculative Beam Extension produces an entire future CoT step for a candidate beam.  
In a standard pipeline, this would trigger two separate verifier calls across iterations, one for the current step and another for the speculative step in the next.
We address this with \textbf{LookAhead Verification}, which exploits the verification locality created by speculation.  
Instead of verifying the two steps separately, we concatenate the output of the current step with the speculative step and submit them together as a single verifier request in the current iteration.
If the speculative path is ultimately chosen, this reduces total verifier latency by improving \textbf{KV cache locality}.  
Processing the two adjacent steps as a continuous sequence allows the verifier to reuse the same KV cache, avoiding costly evictions due to limited memory and eliminating the potential need to recompute key–value states in the next iteration.

\subsection{Dynamic Prefix-Aware Scheduling} ~\label{sec:prefix}
\begin{figure}[t]
    \centering
    \includegraphics[width=0.99\linewidth]{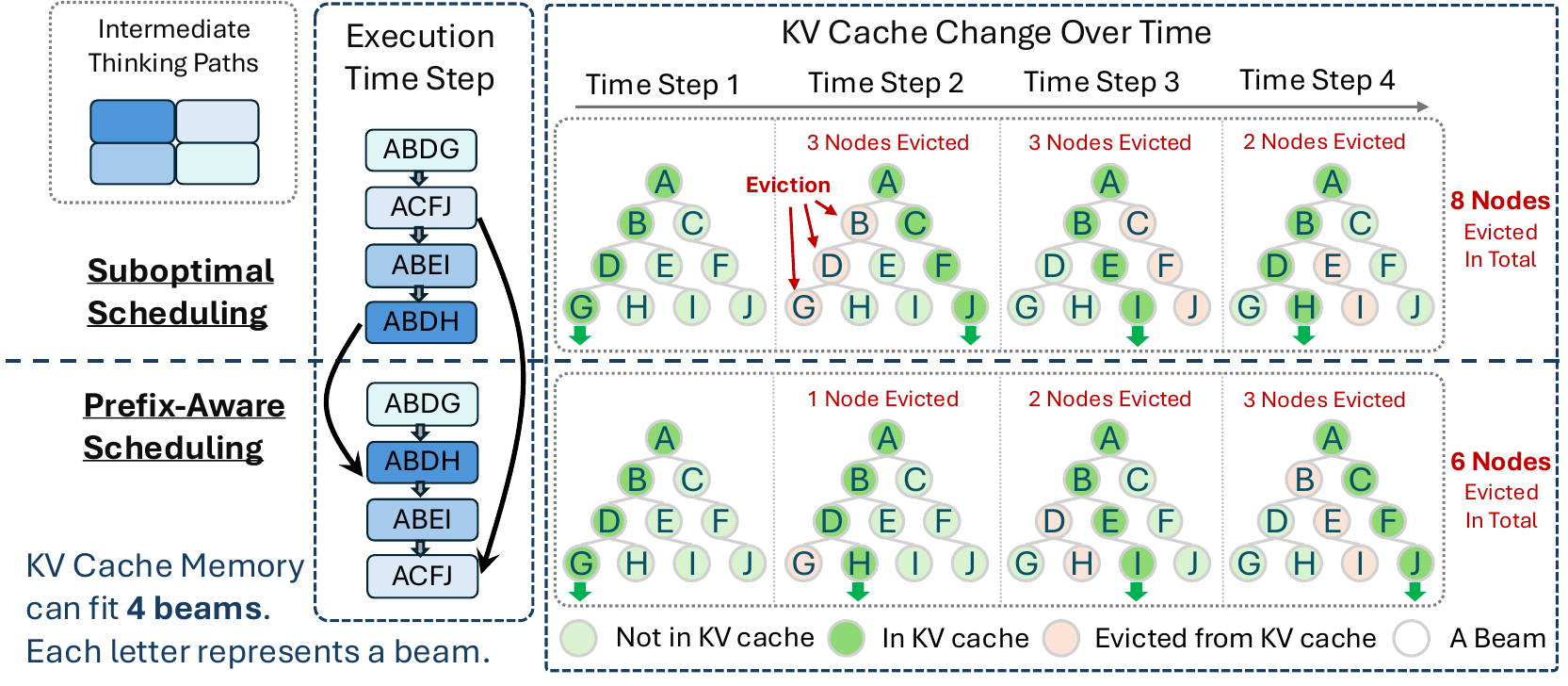}
    \caption{Dynamic Prefix-Aware Scheduling. Example showing how reordering intermediate thinking paths reduces KV cache eviction. Input thinking paths are stored in memory as a Radix Tree. For clarity, the KV cache of newly generated output tokens is omitted in the diagram.
    }
    \label{fig:prefix_diagram}
\end{figure}

Based on the motivation to exploit the unique temporal locality in the generation phase of verifier-guided TTS, our objective is to minimize KV cache evictions over time by intelligently ordering computations using Dynamic Prefix-Aware Scheduling (\figref{fig:prefix_diagram}). 
We first frame this as an optimization problem.
At each iteration of the generation process, the scheduler receives a list of active reasoning paths, or CoTs, where each CoT is a sequence of beams. A schedule, $S$, determines the processing order for this list of CoTs. 
Given a constrained KV cache memory budget, the ordered list is partitioned into batches. Each batch is represented as a radix tree (Trie), $T_i$, which is the largest possible group of consecutively scheduled CoTs that can fit into memory. Within a Trie, each node represents a unique beam.

We model the cost of KV cache eviction when switching from processing Trie $T_i$ to $T_{i+1}$ as the number of old nodes that must be evicted from memory. The total eviction cost is the sum of these costs over the entire schedule:
\[
\text{Cost} = \sum_{i} (\text{Nodes}(T_i) - P(T_i, T_{i+1}))
\]
Here, $\text{Nodes}(T_i)$ is the node count of Trie $T_i$, and $P(T_i, T_{i+1})$ is the size of the shared prefix (i.e., the number of common nodes) between the two consecutive Tries. 
To facilitate our analysis, we assume that the sum $\sum_{i} \text{Nodes}(T_i)$ is constant. A complete list of assumptions is provided in Appendix~\ref{app:prefix_assumptions}.
Minimizing the eviction cost is equivalent to maximizing the sum of shared prefixes. 
Therefore, the optimization problem is to find the schedule $S^*$ that achieves this:
\[
S^* = \underset{S}{\text{argmax}} \left( \sum_{i} P(T_i, T_{i+1}) \right)
\]
We solve this optimization problem using a greedy approach. Given the set $Q$ of CoTs to be scheduled, the following scheduling invariant is maintained:
\[
T_{k+1} = \underset{c_i \in Q}{\text{argmax}} \, P(c_k, c_i)
\]
In practice, the local maximization strategy serves as an effective heuristic, often approaching the global optimum empirically. We establish the local optimality of this strategy under certain assumptions. A formal proof, based on a pairwise interchange argument, is provided in Appendix~\ref{app:prefix_proof}.
We implement this greedy approach efficiently by grouping beams spawned from the same parent beam within the scheduling queue, while preserving the relative order of the parent beams across iterations.

\subsection{Asymmetric Multi-Model Memory Allocation}
\label{ssec:memory_allocation}

\subsubsection{Roofline-Guided KV Allocation} ~\label{sec:roofline_kv}
\begin{figure}[t]
    \centering
    \includegraphics[width=0.99\linewidth]{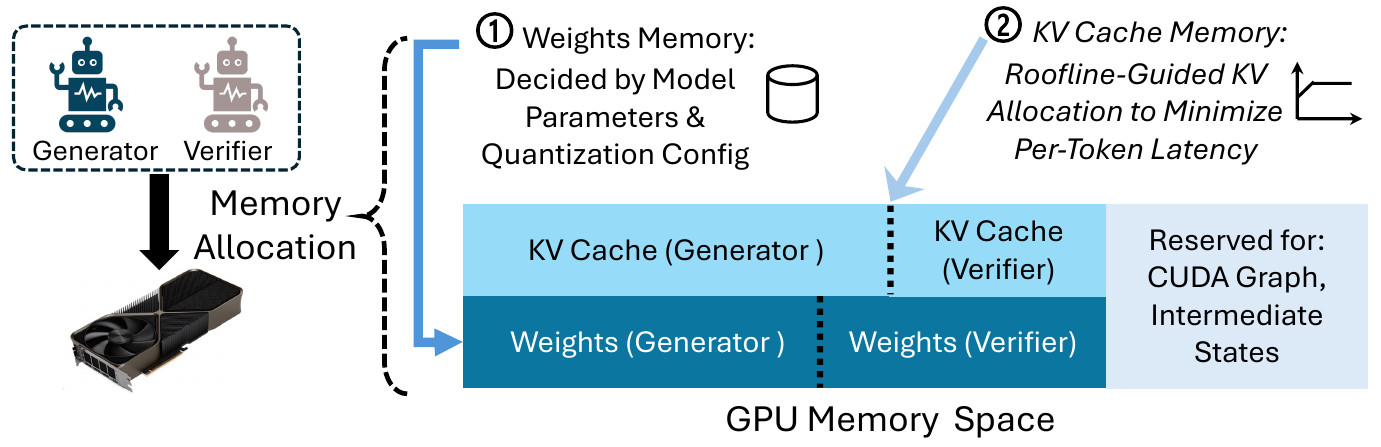}
    \caption{Asymmetric Multi-Model Memory Allocation.
    }
    \label{fig:memory_alloc_diagram}
\end{figure}

As established in our motivation (\secref{motivation:memory}), the available KV cache memory greatly affects system throughput. 
Statically partitioning memory between the verifier and generator is often suboptimal due to their distinct compute patterns. 
We therefore propose a roofline-guided KV allocation strategy that balances the KV cache between the generator and verifier to maximize overall system throughput (\figref{fig:memory_alloc_diagram}).

\paragraph{Formulation.}
Our goal is to find the optimal batch sizes for the prefill (verifier) stage, $B_{\mathrm{pre}}$, and the decoding (generator) stage, $B_{\mathrm{dec}}$, that minimize the total execution time, $T_{\mathrm{tot}}$, for a workload of $N$ requests. We define the total time $T_\mathrm{tot}$ as the sum of the time spent in each stage:
\[
T_{\mathrm{tot}}=
\underbrace{\Big\lceil\tfrac{N}{B_{\mathrm{pre}}}\Big\rceil\,T_{\mathrm{roof}}^{\mathrm{pre}}(B_{\mathrm{pre}},S)}_{\text{Total Prefill/Verifier Time}}
\;+\;
\underbrace{\Big\lceil\tfrac{N}{B_{\mathrm{dec}}}\Big\rceil\,S_{\mathrm{dec}}\,
T_{\mathrm{roof}}^{\mathrm{dec}}(B_{\mathrm{dec}},\bar S_{\mathrm{cache}})}_{\text{Total Decoding/Generator Time}},
\]
where $S$ is the input sequence length for the verifier, $S_{\mathrm{dec}}$ is the generation length for the generator, and $\bar{S}{\mathrm{cache}}$ is the average KV cache length during decoding ( $\approx S_{\mathrm{dec}}/2$).
The term $\lceil\tfrac{N}{B}\rceil$ calculates the number of batches required to process all $N$ requests. For the decoding stage, the per-token generation time is multiplied by the decoding horizon $S_{\mathrm{dec}}$.

This optimization is subject to the total KV cache memory budget, $M$:
\[
B_{\mathrm{pre}}\!\cdot\!\mathrm{KVBytes}(1,S)\;+\;
B_{\mathrm{dec}}\!\cdot\!\mathrm{KVBytes}(1,S_{\mathrm{dec}})\;\le\;M.
\]
The latency for a single batch in each stage, $T_{\mathrm{roof}}$, is estimated using a standard Roofline model. This model defines latency as the maximum of the time constrained by compute or by memory bandwidth:
\[
T_{\mathrm{roof}}=\max\!\left(\frac{\mathrm{FLOPs}}{P},\;\frac{\mathrm{Bytes}}{\mathrm{BW}}\right),
\]
where $P$ is the device's peak compute and $\mathrm{BW}$ its peak memory bandwidth, per hardware specification.

\paragraph{Search Algorithm.}
Since the objective function $T_{\mathrm{tot}}$ is not necessarily convex, we employ a simple and fast linear search that is guaranteed to find the global optimum. A key insight is that since stage latency monotonically decreases with more memory, the optimal allocation will always lie on the \emph{boundary} of the memory constraint, fully utilizing the available budget $M$.

\begin{figure}
    \centering
    \includegraphics[width=0.99\linewidth]{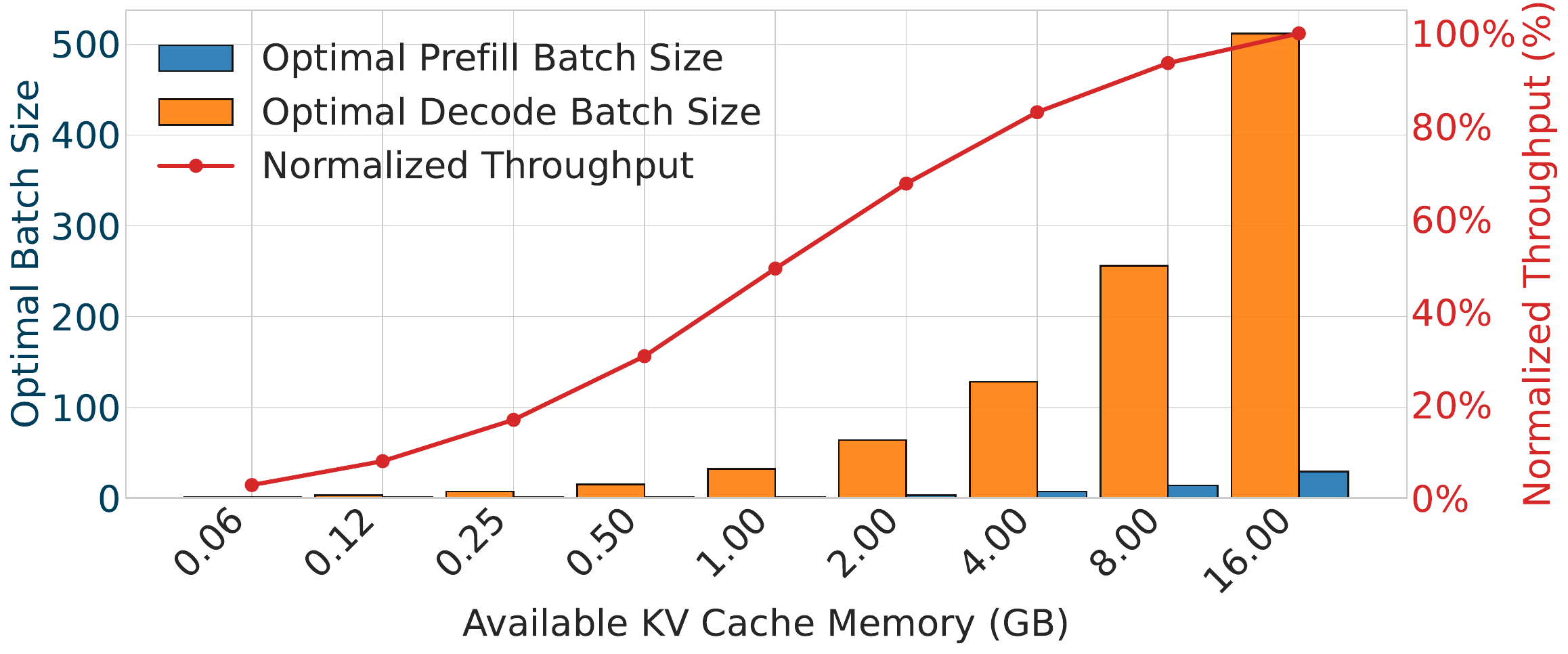}
    \caption{\textbf{Roofline-Guided KV Allocation.} Bars: optimal prefill/decoding batch sizes; line: normalized throughput (higher is better).
    }
    \label{fig:cache_allocation}
\end{figure}

Our search algorithm therefore iterates through all feasible integer values for the prefill batch size, $B_{\mathrm{pre}}$. For each candidate $B_{\mathrm{pre}}$, we calculate the maximum possible decoding batch size, $B_{\mathrm{dec}}$, that satisfies the memory constraint:
\begin{align}
B_{\mathrm{dec}}
&=\Big\lfloor
\tfrac{M - B_{\mathrm{pre}}\mathrm{KVBytes}(1,S)}
{\mathrm{KVBytes}(1,S_{\mathrm{dec}})}
\Big\rfloor.
\end{align}
We then evaluate $T_{\mathrm{tot}}$ for this $(B_{\mathrm{pre}}, B_{\mathrm{dec}})$ pair and record the pair that yields the minimum total time. Because the decoding stage is typically more sensitive to memory, any ties are resolved in favor of a larger $B_{\mathrm{dec}}$. This entire search process is computationally trivial, averaging \textbf{$<1$\,ms} on a single CPU thread, and thus introduces negligible overhead.
\figref{fig:cache_allocation} shows an example resulting policy.
At run-time, the Roofline-Guided KV Allocation policy is dynamically invoked upon system state changes to quickly adapt the verifier and generator batch sizes.

\subsubsection{Extended Search Space with Offloading}
The optimization space can be extended with an offloading strategy for cases where GPU memory $M$ is extremely constrained. 
Here, the KV cache of the inactive model is offloaded to CPU memory, enabling a single model to fully utilize the GPU cache space and relaxing the coupled constraint into two independent ones:
\[
B_{\mathrm{pre}}\!\cdot\!\mathrm{KVBytes}(1,S) \le M, 
\quad 
B_{\mathrm{dec}}\!\cdot\!\mathrm{KVBytes}(1,S_{\mathrm{dec}}) \le M.
\]
This incurs a transfer overhead $T_{\text{overhead}}^{\text{offload}}$.
The system then selects the lower-latency strategy: 
\textit{i)} the optimal execution time $T_{\mathrm{tot}}$ from allocation search under the original constraint, or 
\textit{ii)} the offloading time $T_{\mathrm{tot}}^{\text{offload}} + T_{\text{overhead}}^{\text{offload}}$, 
where $T_{\mathrm{tot}}^{\text{offload}}$ is computed from the maximum batch sizes allowed by the relaxed constraints. 
This dual-strategy policy lets \fasttts{} always pick the better option.

\section{Implementation}

\fasttts{} is implemented in \textasciitilde6,500 lines of Python on top of vLLM (v0.9.2), operating the generator and verifier in separate worker processes via Python's multiprocessing library. We extend the core \texttt{LLMEngine} of vLLM to implement our two-phase, preemptive scheduling policy, which dynamically switches between continuous batching and \textbf{Speculative Beam Extension} based on the request queue status. 
For the generator, we extend the default scheduler with our \textbf{Dynamic Prefix-Aware Scheduling} that implements a greedy heuristic to group beams from the same parent, maximizing KV cache reuse. 
Our \textbf{Asymmetric Multi-Model Memory Allocation} policy is managed by a lightweight searcher that is invoked dynamically to determine the partition of the KV cache between workers. 
The system exposes a configurable interface for various TTS strategies and hyperparameters. 

\begin{figure}[t]
    \centering
    \includegraphics[width=\linewidth]{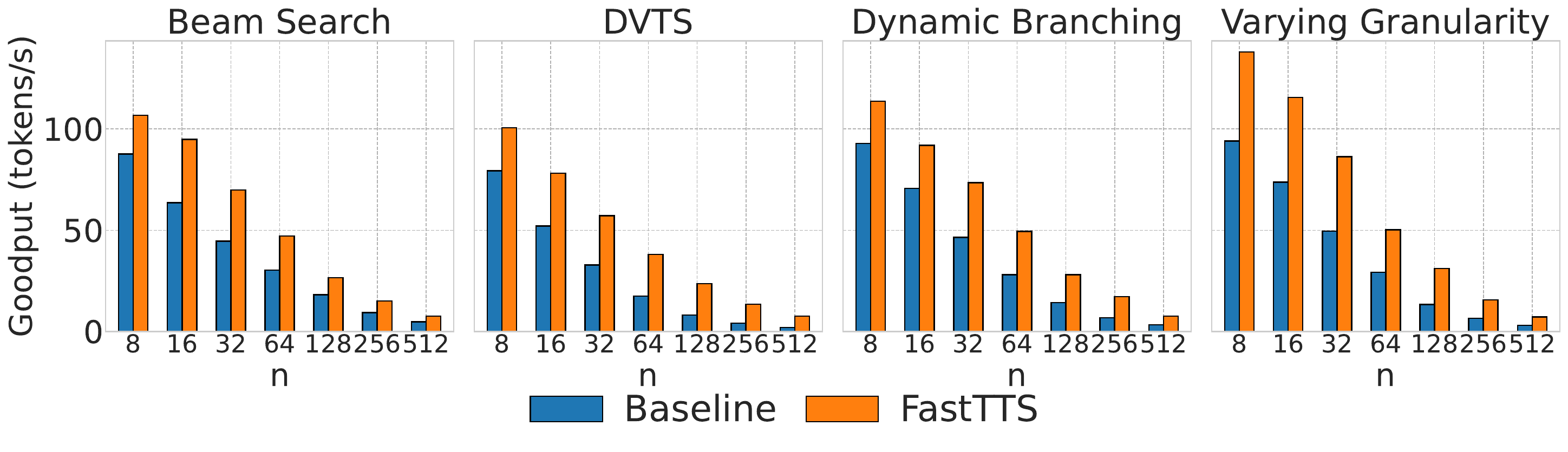}
    \caption{Precise Goodput improvement of \fasttts{} over the vLLM baseline across different search algorithm variants.
    Experiments use the 1.5B+1.5B configuration on AIME. In dynamic branching, each beam branches proportionally to its verifier score; in varying granularity, the maximum step length is 64 tokens for the first 3 steps and 2048 thereafter.
    }
    \label{fig:search_variants_goodput}
\end{figure}

\begin{figure*}[t]
    \centering
    \includegraphics[width=0.99\linewidth]{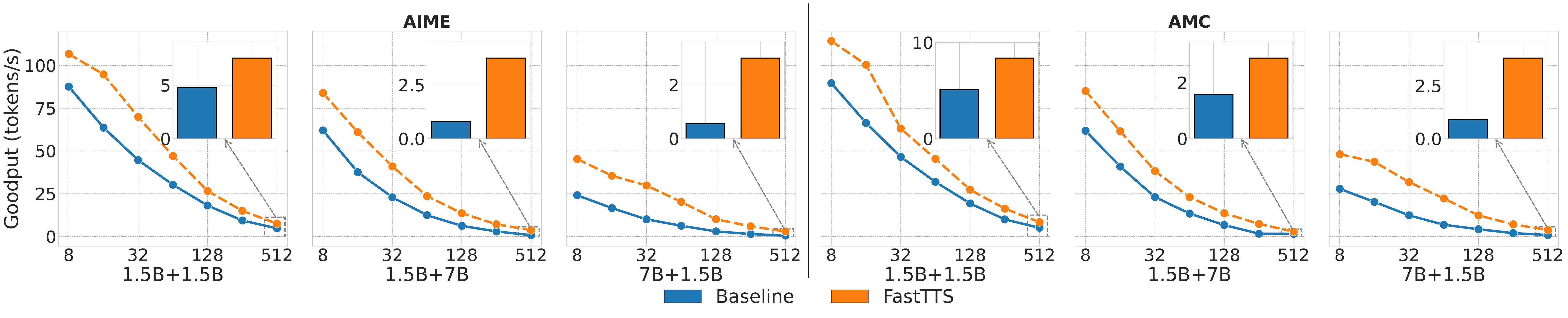}
    \caption{\textbf{\fasttts{} Goodput Improvement}. The $x$-axis represents the number of beams (\textit{n}).
    }
    \label{fig:exp_main}
\end{figure*}

\begin{figure*}
    \centering
    \includegraphics[width=0.99\linewidth]{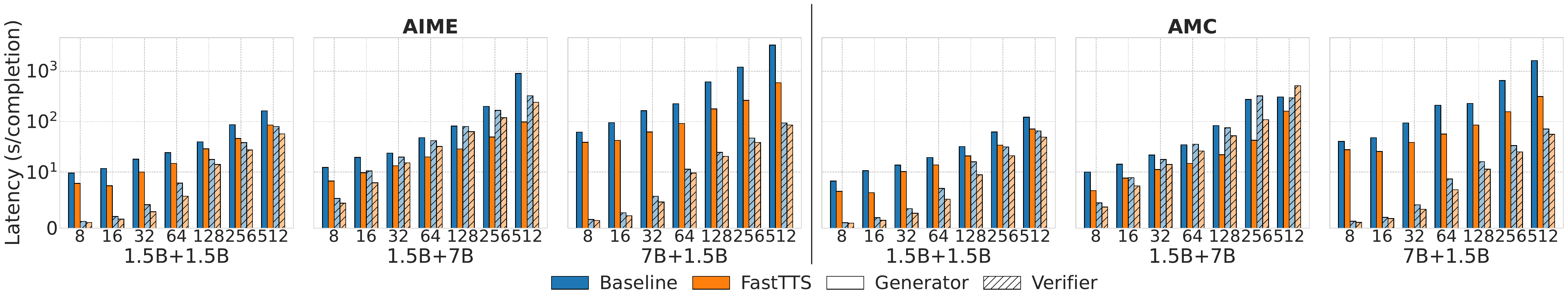}
    \caption{\textbf{Completion Latency Improvement}. The average end-to-end latency for the completion of a single request. The search process is terminated when all active beams reach the end. The $x$-axis represents the number of beams.
    }
    \label{fig:exp_latency}
\end{figure*}

\section{Evaluation}\label{sec:exp}
\subsection{Experimental Setup} \label{exp:setup}

\noindent\textbf{Platform.}
All experiments are conducted on a single NVIDIA GeForce RTX 4090 GPU (24 GB VRAM), representing a typical edge device. It is equipped with an Intel Xeon Silver 4310 CPU @ 2.10 GHz. The software stack includes CUDA Toolkit 12.4 with its corresponding versions of Nsight Systems and Nsight Compute, PyTorch 2.7.0, and Python 3.11.

\noindent\textbf{Models.}
To assess \fasttts{} under diverse workloads, we evaluate three generator-verifier configurations designed to stress different system aspects, following common practice in prior work~\cite{snell2024scaling, chen2025rethinking, beeching2024scalingtesttimecompute, liu2025can}. We test a verifier-heavy setting (\textbf{1.5B+7B}: Qwen2.5-Math-1.5B generator with a Math-Shepherd-Mistral-7B verifier) and a generator-heavy setting (\textbf{7B+1.5B}: Qwen2.5-Math-7B generator with a Skywork-o1-Open-PRM-1.5B verifier), both allocate 90\% of GPU memory to test throughput limits. To simulate a highly resource-limited environment, we also test a memory-constrained setting (\textbf{1.5B+1.5B}: a 1.5B generator and verifier)~\cite{yang2024qwen2, skyworkopeno12024, wang2023math}, restricting it to 40\% of GPU memory.

\noindent\textbf{Datasets.}
We evaluate on two common mathematical reasoning benchmarks~\cite{yang2024qwen2, shao2024deepseekmath} of varying difficulty to assess performance under diverse and complex workloads:
\begin{itemize}[leftmargin=*,topsep=0pt,itemsep=0pt,parsep=0pt]
    \item \textbf{AIME2024}~\cite{maa2024aime}: A challenging dataset from the American Invitational Mathematics Examination. 
    \item \textbf{AMC2023}~\cite{balunovic_srimatharena_2025}: A dataset from the American Mathematics Competitions, which presents a broader range of difficulty. 
\end{itemize}
For experiments, we use the test sets of these benchmarks with a batch size of 1 to reflect interactive edge scenarios.

\noindent\textbf{Baseline Implementation.}
Our baseline system is built on top of the widely-used \textbf{vLLM} framework (version 0.9.2). We implement a standard verifier-guided test-time search, running the generator and verifier as separate vLLM instances, with remaining details following Hugging Face’s official \textit{search-and-learn} implementation~\cite{beeching2024scalingtesttimecompute}. 
This baseline represents a naive but robust implementation of TTS, against which we compare the performance gains achieved by the optimizations in \fasttts{}.

\noindent\textbf{Metrics.} 
To provide a comprehensive evaluation of system performance for TTS, we use the following metrics:
\begin{itemize}[leftmargin=*,topsep=0pt,itemsep=0pt,parsep=0pt]
    \item \textbf{Precise Goodput}: Standard goodput metrics are often insufficient for TTS tasks. To fairly evaluate system efficiency, we propose a metric termed \textbf{Precise Goodput}\footnote{We used Precise Goodput and Goodput interchangeably in this paper.}, defined as:
    \[ \text{Precise Goodput} := \frac{\text{Average token length per beam}}{\text{Average beam completion time}} \]
    This metric is designed to be robust against several sources of evaluation unfairness. Using the average completion time and token length across all beams prevents the metric from being affected by a single slow reasoning path or being artificially inflated by a large number of finally collected paths. Furthermore, it provides a true measure of generation efficiency, unaffected by the copying of text during branching.
    \item \textbf{Completion Time}: We measure the average end-to-end time taken per completion for a problem.
\end{itemize}

\subsection{End-to-End Performance Improvement}

We first evaluate the end-to-end performance of \fasttts{} against the vLLM baseline across a diverse set of popular test-time search algorithms. As shown in \figref{fig:search_variants_goodput}, \fasttts{} consistently and significantly improves precise goodput over the baseline implementation across all evaluated search methods.  
The goodput improvement ranges from 1.2$\times$ to 3.9$\times$. 
DVTS, Dynamic Branching, and Varying Granularity are fundamentally variants of the core beam search algorithm. As beam search represents the most common and foundational use case, we focus the remainder of our evaluation on this representative search method.

\noindent\textbf{Precise Goodput.}
\figref{fig:exp_main} shows that \fasttts{} consistently and significantly improves system goodput over the vLLM baseline across all tested scenarios. 
For all three model configurations (1.5B+1.5B, 1.5B+7B, and 7B+1.5B) and number of beams (\textit{n}) values from 8 to 512, \fasttts{} achieves an average goodput improvement of \boldmath{$2.2\times$}, ranging from \boldmath{$1.2\times$} \textbf{to} \boldmath{$5.4\times$}. 
These substantial gains stem from our synergistic optimizations: Speculative Beam Extension enhances GPU utilization, while Asymmetric Multi-Model Memory Allocation and Dynamic Prefix-Aware Scheduling improve the efficiency of KV cache management.
The relative goodput improvement becomes more pronounced at larger values of \textit{n}, peaking at 
5.4$\times$ for the 7B+1.5B configuration at \textit{n}=512 on AIME. This trend holds for all model pairs, as a larger search budget (\textit{n}) creates more diverse reasoning paths, further exacerbating hardware underutilization and KV cache pressure---the very issues our optimizations address.

\noindent\textbf{Completion Latency.}
Beyond improving goodput, \fasttts{} also delivers substantial reductions in end-to-end completion latency. \figref{fig:exp_latency} shows that \fasttts{} achieves an average latency reduction of \textbf{38\% to 68\%} across all configurations and \textit{n} values compared to the vLLM baseline.

The latency breakdown within \fasttts{} reveals the distinct performance characteristics of each model configuration. In the 7B+1.5B configuration, generator latency (unfilled portion) is the dominant cost. Conversely, in the 1.5B+7B configuration, the larger 7B verifier model contributes a substantial portion of the total latency, becoming nearly on par with the generator as \textit{n} increases.

\fasttts{} effectively reduces both the generation and verification components of latency. On average, it reduces verifier latency by \textbf{75\% to 85\%} and reduces generation latency by \textbf{36\% to 66\%} across all \textit{n} values. The dramatic reduction in verifier latency is primarily driven by our \textit{LookAhead Verification} technique, which enhances computational locality by pre-verifying tokens. The substantial decrease in generator latency is achieved through the combined effects of our other optimizations: Asymmetric Multi-Model Memory Allocation and Dynamic Prefix-Aware Scheduling enhance KV cache efficiency, while Speculative Beam Extension hides straggler latency by utilizing idle GPU cycles.
We note one exception where verifier latency slightly increases (at \textit{n}=512 for the 1.5B+7B model on AMC), a direct trade-off from our memory allocator prioritizing the heavily-loaded generator by reducing the verifier's KV cache capacity.

\subsection{Algorithm Performance}

\begin{figure}[t]
    \centering
    \begin{subfigure}{0.495\linewidth}
        \centering
         \includegraphics[width=\linewidth]{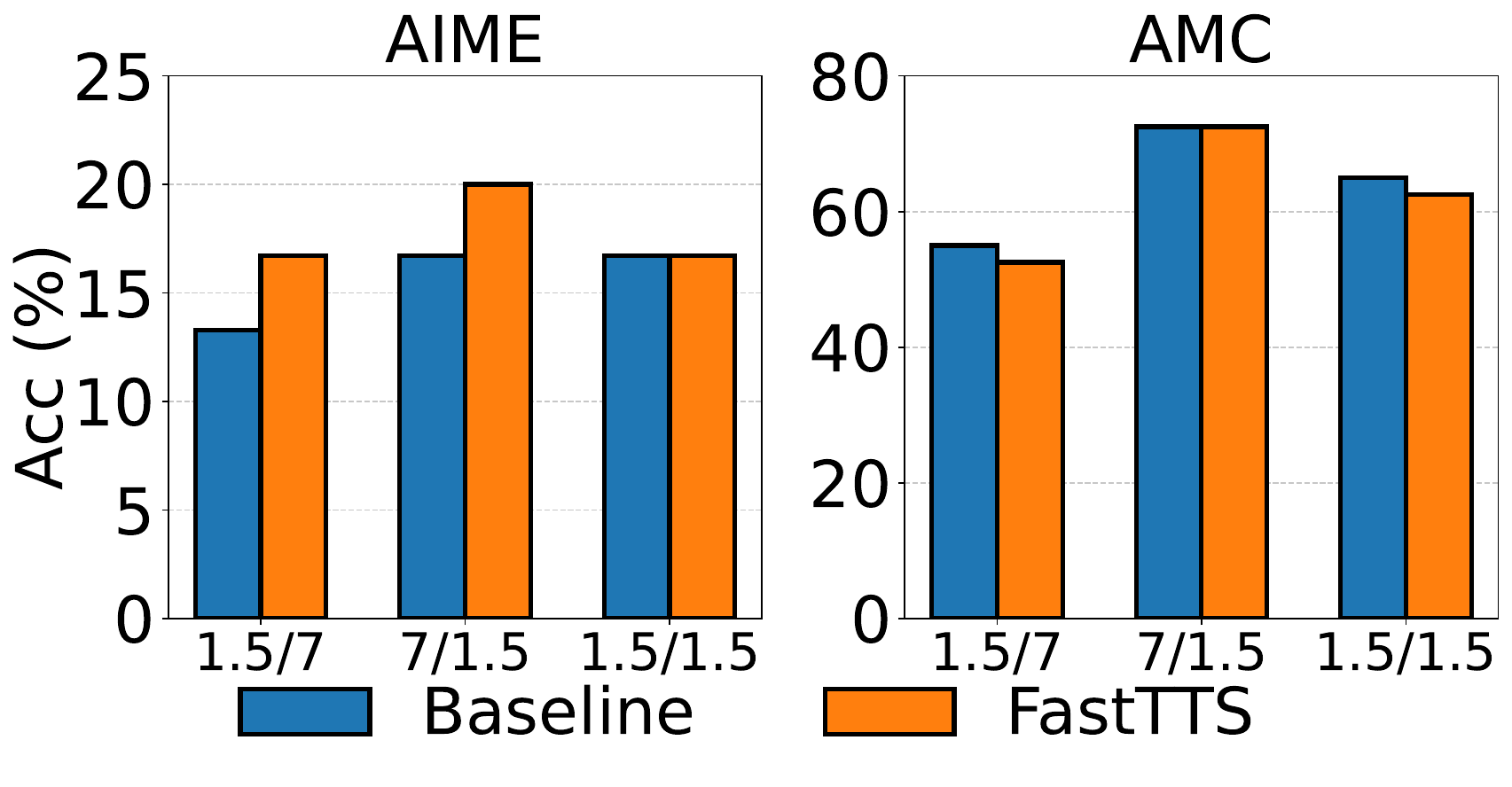}
        \caption{Top-1 Accuracy with n=512.  }
    \label{fig:top1_acc}
    \end{subfigure}
    \hfill
    \begin{subfigure}{0.495\linewidth}
        \centering
        \includegraphics[width=\linewidth]{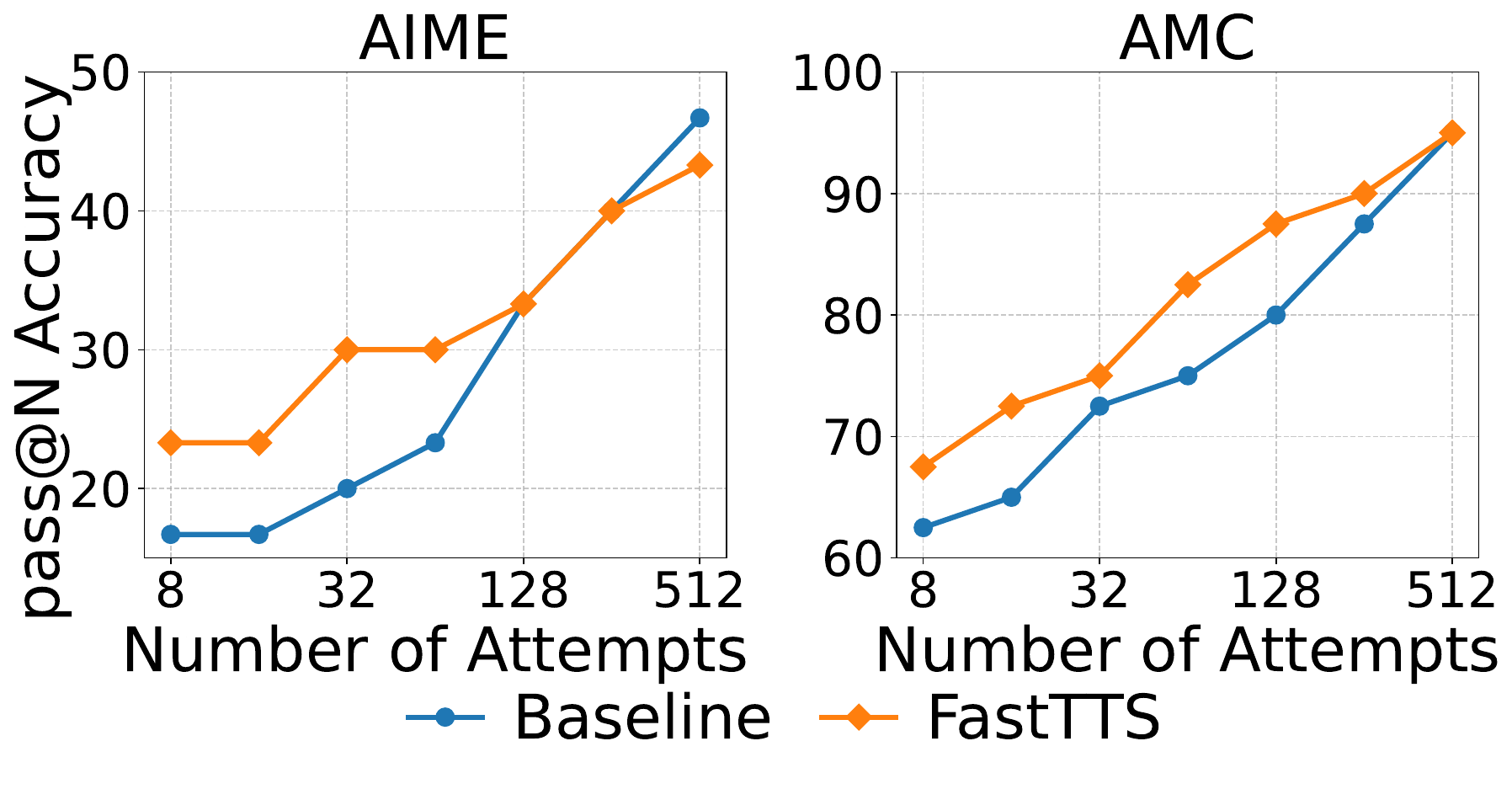}
        \caption{Pass@N accuracy.}
        \label{fig:pass_n_acc}
    \end{subfigure}
    \caption{Algorithm accuracy (e.g., 1.5/7 for 1.5B+7B). 
    }
    \label{fig:exp_acc}
\end{figure}



We evaluate the impact of our system optimizations on the quality of the generated solutions from two perspectives. For a practical assessment, we report \textbf{Top-1 accuracy}, where the final answer is selected from the generated candidates using majority voting. To better understand the quality distribution of all generated solutions and the capability of the search algorithm, we also report \textbf{Pass@N accuracy}. This metric measures the success rate where at least one correct answer is found within a set of N generated solutions. For ranking, the N candidates are selected based on their verifier score. 

While \fasttts{} is designed to guarantee algorithmic equivalence with the baseline, minor variations in output can occur since our scheduling optimizations may alter the sampling order. We now analyze these effects.

\noindent\textbf{Top-1 Accuracy.}
As shown in \figref{fig:top1_acc}, the Top-1 accuracy of \fasttts{} is highly competitive with the baseline. On the more challenging AIME dataset, \fasttts{} consistently matches or slightly improves accuracy, likely because its speculative execution focuses computation on the more promising reasoning paths. In general, both perform comparably, confirming the algorithm equivalence.

\noindent\textbf{Pass@N Accuracy.}
\figref{fig:pass_n_acc} shows the Pass@N accuracy, providing insight into the search behavior. 
In practice, it matches baseline accuracy at large $N$ but slightly exceeds it at small $N$, likely due to a side scheduler effect: speculative extension can let long straggler beams generate beyond their original CoT length, occasionally improving accuracy.



Ultimately, for practical deployment, the Top-1 accuracy achieved through majority voting is the more indicative measure of a system's real-world utility.

\begin{figure}[t]
    \centering
    \includegraphics[width=\linewidth]{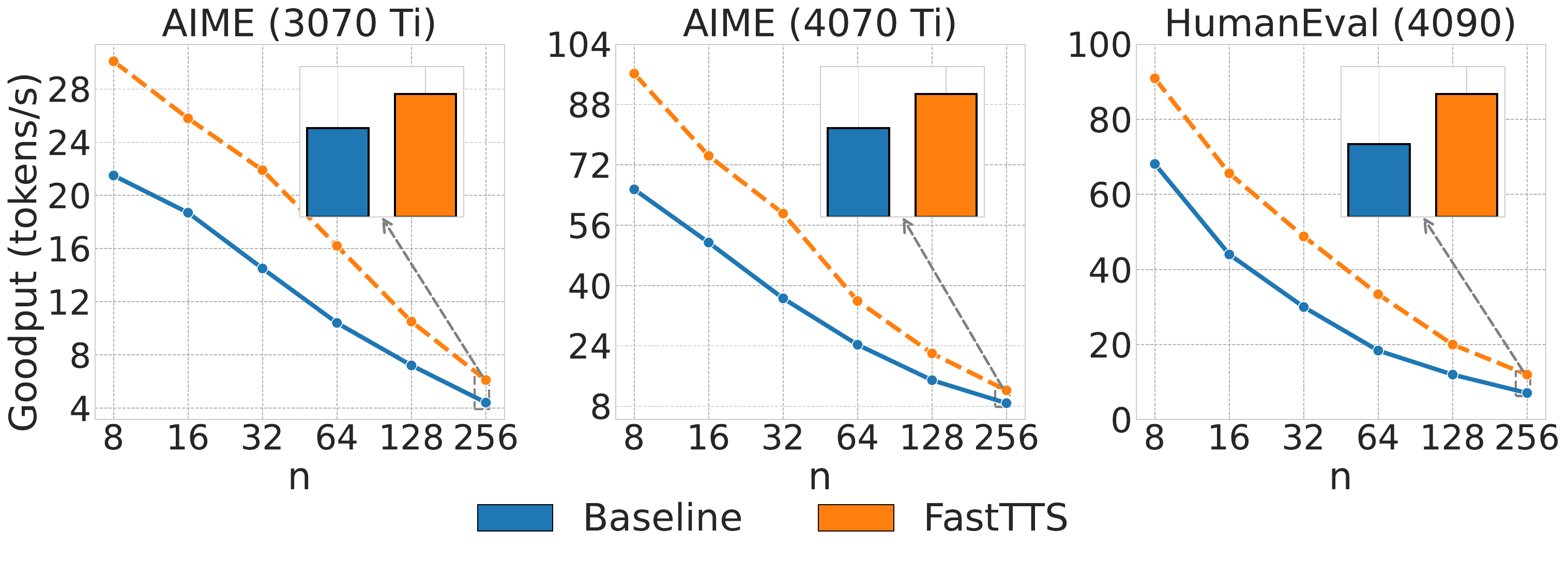}
    \caption{
    Goodput improvements on constrained hardware and coding tasks. Experiments are conducted on NVIDIA RTX 3070 Ti (8GB) and RTX 4070 Ti (12GB) GPUs using the AIME dataset, and on the HumanEval code generation benchmark using an RTX 4090. Offloading is used on the RTX 3070 Ti; as a result, lower absolute goodput is observed.
    }
    \label{fig:more_hardware_benchmarks}
\end{figure}

\subsection{Generality on Hardware and Benchmarks}

To demonstrate the generality of \fasttts{} across more resource-constrained GPUs and diverse tasks, we further extend our evaluation to additional devices and code generation workloads.

\textbf{Constrained Hardware.} We evaluate \fasttts{} on NVIDIA RTX 3070 Ti (8GB) and RTX 4070 Ti (12GB) GPUs. As shown in Fig.~\ref{fig:more_hardware_benchmarks}, our system consistently outperforms the baseline, achieving goodput speedups of \textbf{1.4$\times$--1.6$\times$}. These results indicate that \fasttts{} remains effective on lower-end edge hardware. We note that \fasttts{} is orthogonal to quantization and offloading techniques, which can be incorporated for additional efficiency gains.

\textbf{Broader Benchmarks.} On the HumanEval code generation benchmark, \fasttts{} attains speedups ranging from \textbf{1.3$\times$ to 1.8$\times$} (Fig.~\ref{fig:more_hardware_benchmarks}). This demonstrates that the execution patterns optimized by \fasttts{} transfer effectively to other complex domains, including code generation.

\subsection{Ablation Study}

\subsubsection{Goodput Gain Breakdown}

\begin{figure}
    \centering
    \includegraphics[width=\linewidth]{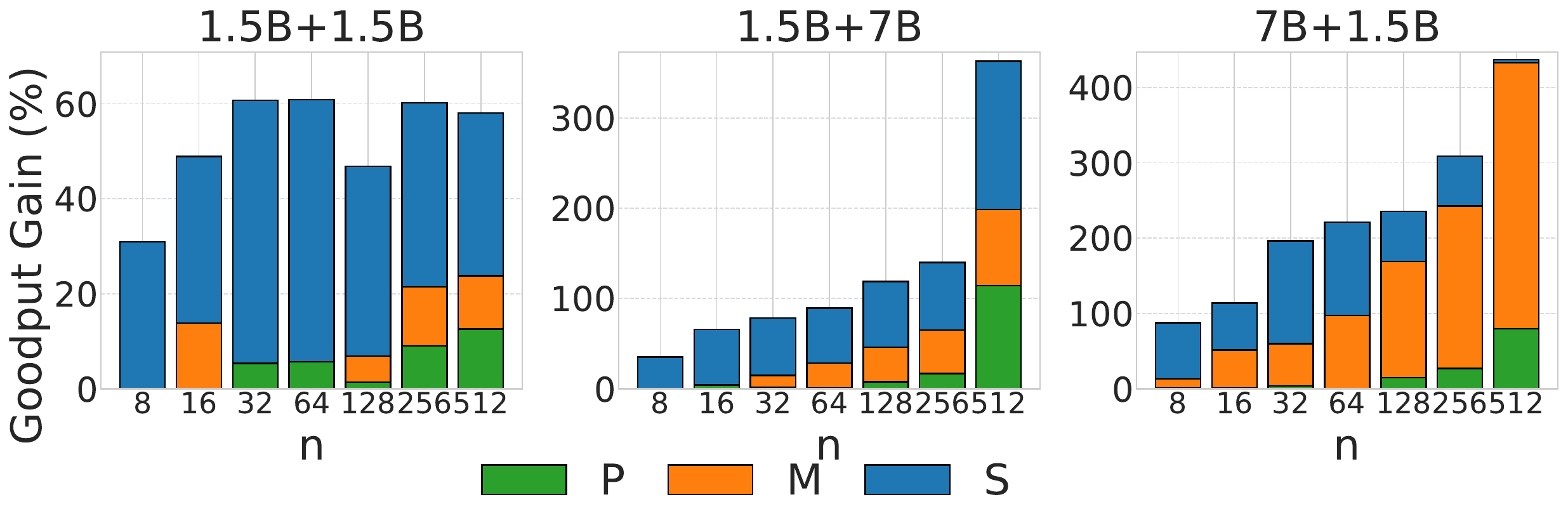}
    \caption{Breakdown of goodput gain from 3 optimizations. The cumulative improvements are shown for Dynamic Prefix-Aware Scheduling (P), Asymmetric Multi-Model Memory Allocation (M), and Speculative Beam Extension (S).
    }
    \label{fig:ablation}
\end{figure}


To understand the individual contribution of each of our proposed optimizations, we conduct an ablation study. The results, shown in \figref{fig:ablation}, break down the cumulative performance gains for all three model configurations: 1.5B+1.5B, 1.5B+7B, and 7B+1.5B.

\noindent\textbf{Dynamic Prefix-Aware Scheduling (P).}
This optimization provides a foundational layer of improvement that becomes more apparent as \textit{n} increases. As shown by the green bars, its gain is most significant in memory-constrained scenarios (e.g., the 1.5B+7B setup), where maximizing prefix reuse is critical. This is intuitive, as a larger number of beams (\textit{n}) leads to a more constrained KV cache where minimizing evictions is paramount.

\noindent\textbf{Asymmetric Multi-Model Memory Allocation (M).}
Adding Asymmetric Multi-Model Memory Allocation on top of Dynamic Prefix-Aware Scheduling delivers additional performance improvement. This component is a major source of improvement across all three model configurations, particularly at larger \textit{n}. This is because under a high compute budget, intelligently partitioning memory between the generator and verifier is crucial to prevent frequent preemptions and costly re-computation for the generator.

\noindent\textbf{Speculative Beam Extension (S).}
Speculative Beam Extension consistently provides a significant, and often the largest, performance improvement. This technique offers a substantial goodput gain across almost all scenarios. The improvement is most pronounced when more KV cache memory is available for parallel speculation, such as in the 1.5B+1.5B and 1.5B+7B configurations. By effectively hiding the latency of straggler beams, this optimization improves goodput.

\subsubsection{In-depth Study of Speculative Beam Extension}

\begin{figure}[t]
    \centering
    \begin{subfigure}{0.45\linewidth}
        \centering
        \includegraphics[width=\linewidth]{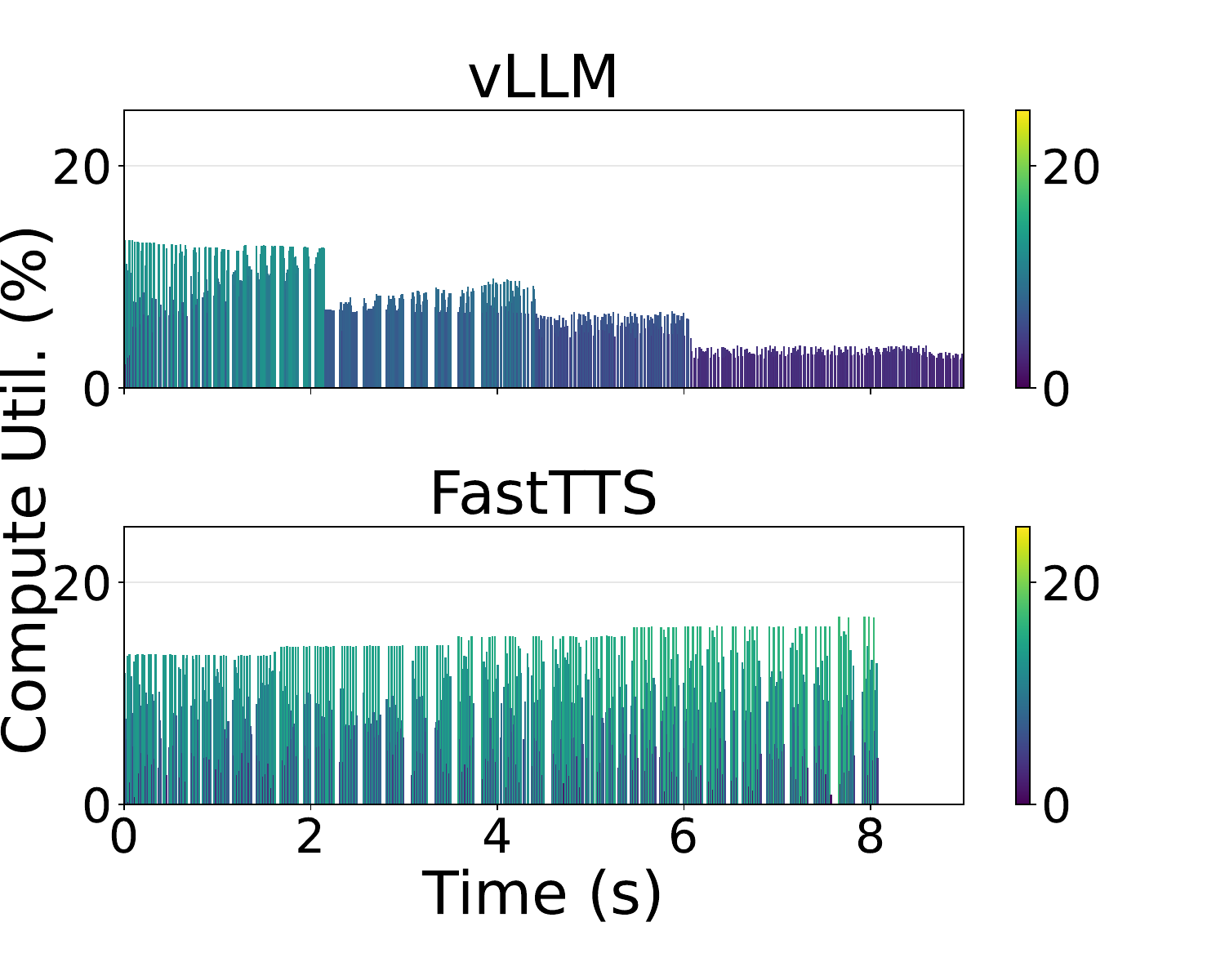}
    \end{subfigure}
    \hfill
    \begin{subfigure}{0.54\linewidth}
        \centering
        \includegraphics[width=\linewidth]{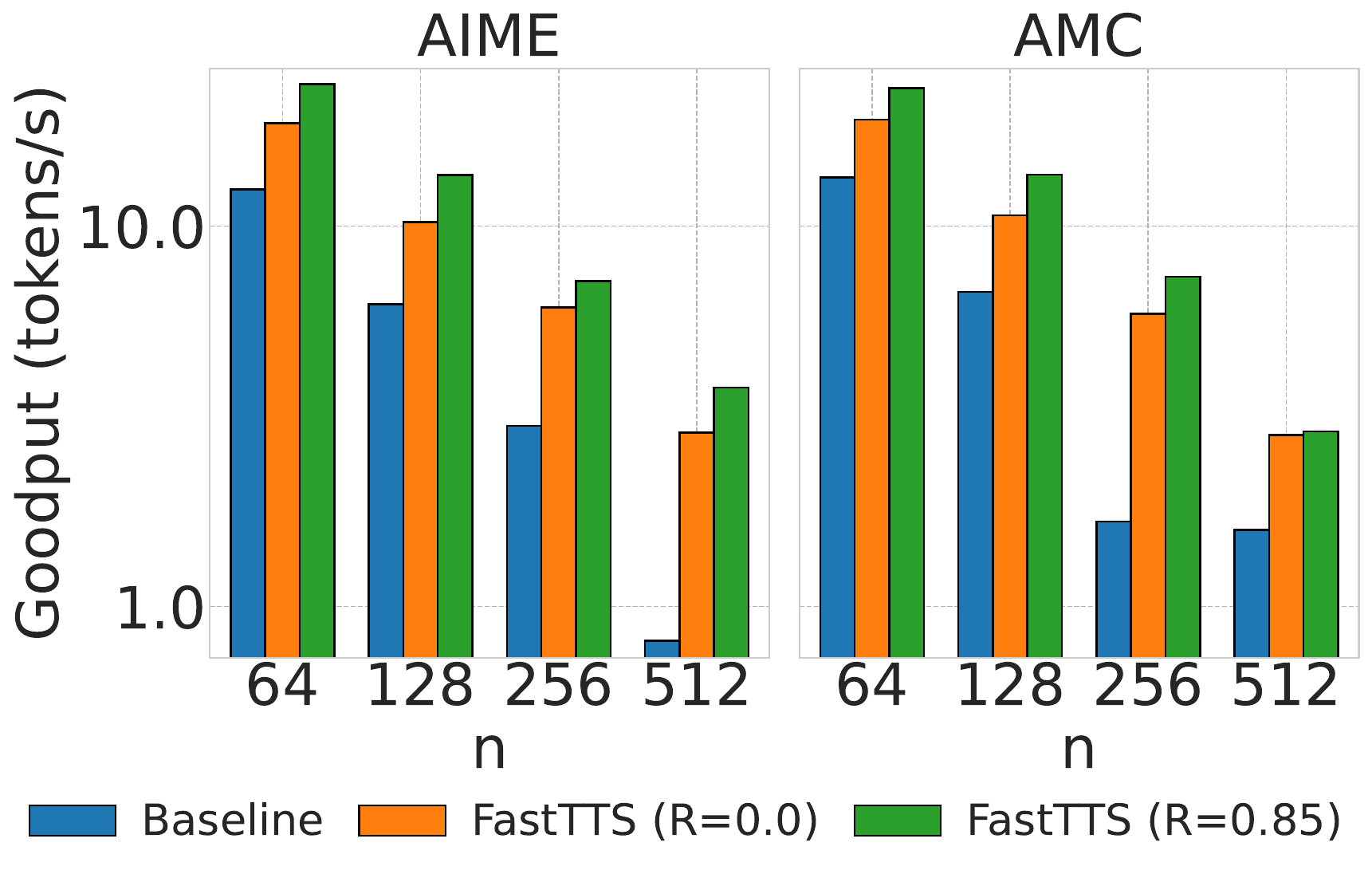}
    \end{subfigure}
    \caption{
    Spec Beam Extension Ablation. Left: Compute Utilization across time in 1 iteration. Right: The impact of the speculative truncation ratio (R) on Goodput.
    }
    \label{fig:spec_beam_ablation}
\end{figure}

\begin{figure}[t]
    \centering
    \begin{subfigure}{0.495\linewidth}
        \centering
        \includegraphics[width=\linewidth]{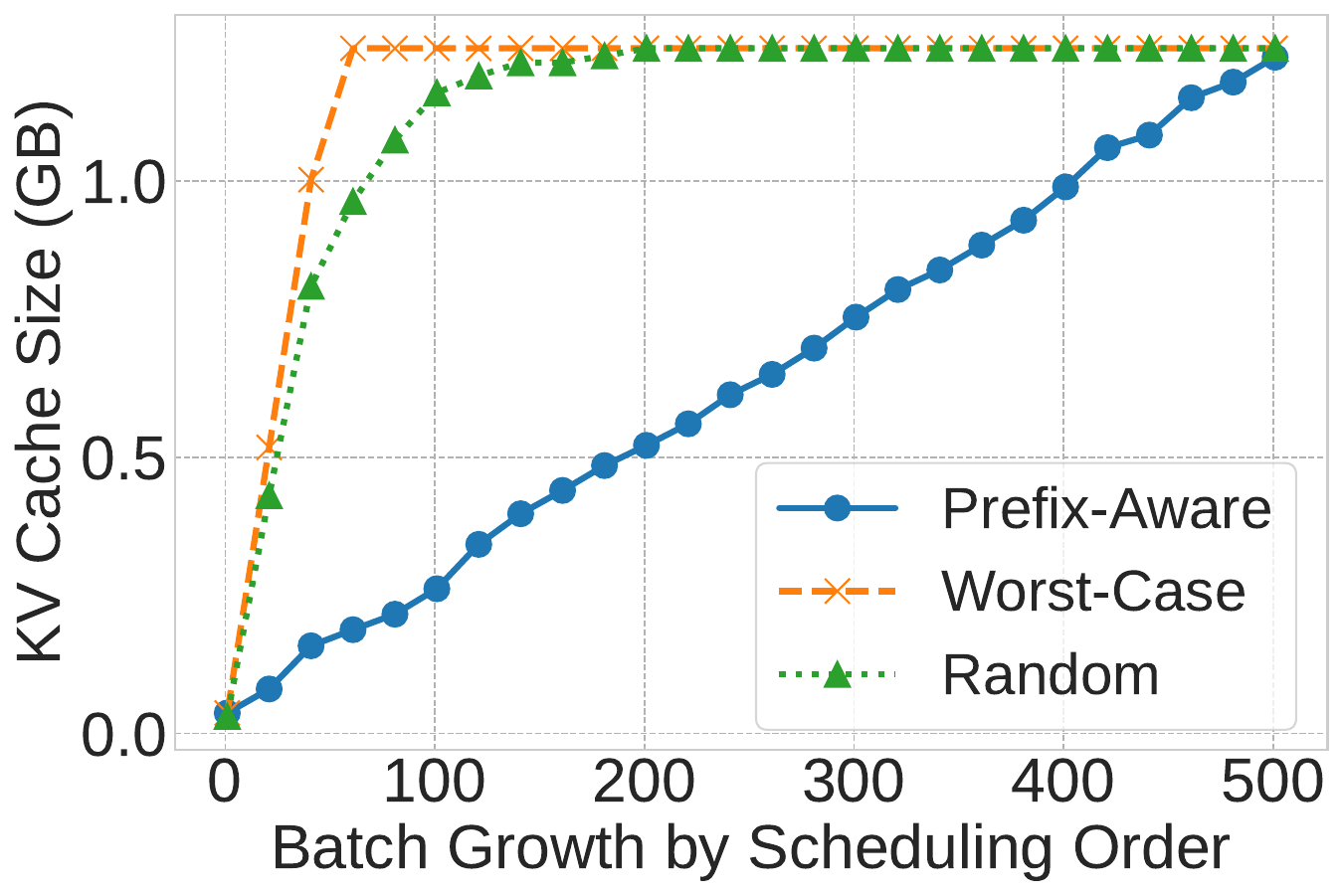}
    \end{subfigure}
    \hfill
    \begin{subfigure}{0.495\linewidth}
        \centering
        \includegraphics[width=\linewidth]{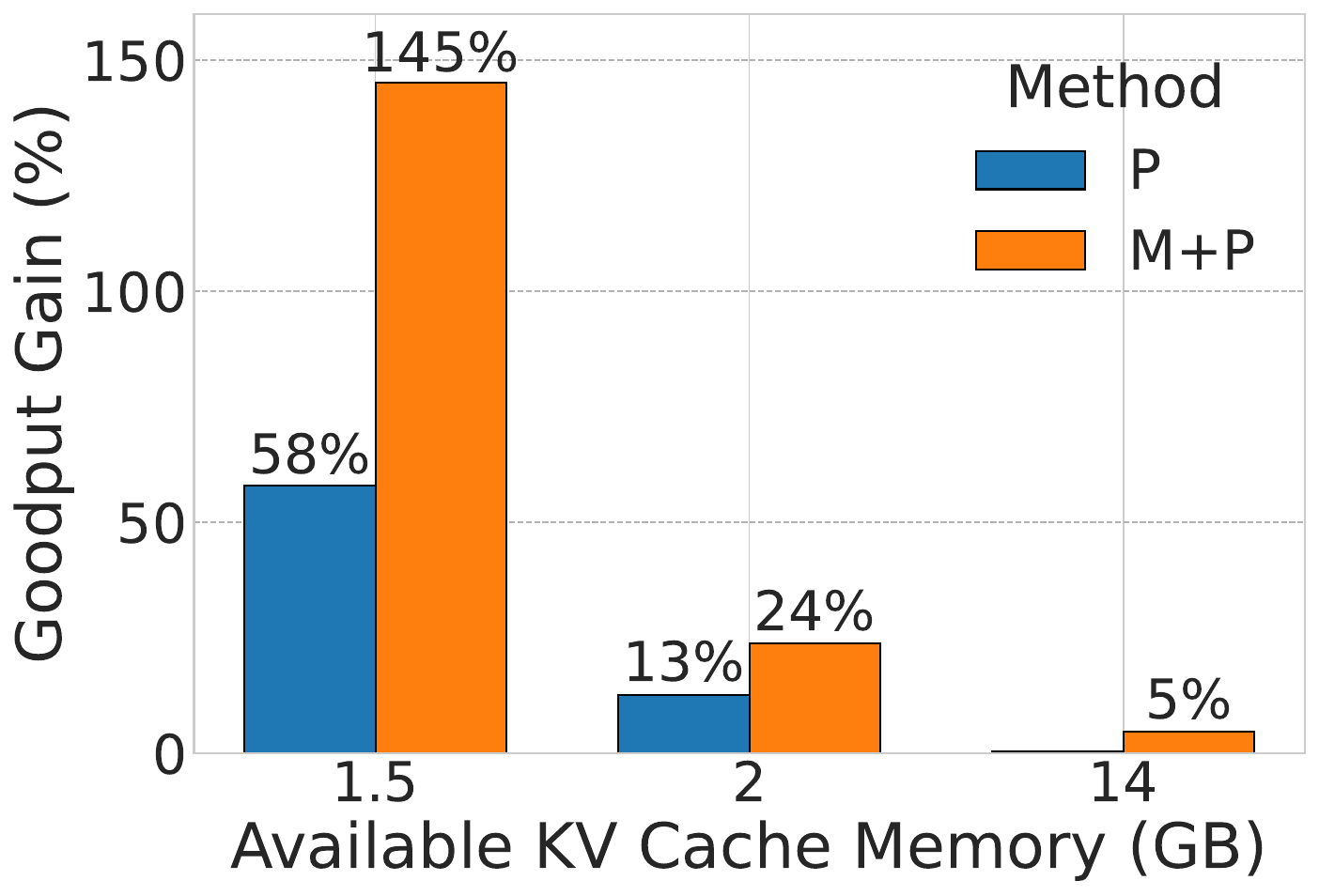}
    \end{subfigure}
    \caption{
    Left: Effectiveness of Dynamic Prefix-Aware Scheduling, using traces from the final TTS iteration (1.5B+1.5B, AIME). The vLLM baseline uses random scheduling. Right: Impact of Memory Availability on Optimization Gains. The chart shows the goodput gain over baselines.
    }
    \label{fig:exp_prefix_memory}
\end{figure}
As shown in \figref{fig:spec_beam_ablation} (left), Speculative Beam Extension improves hardware utilization. While the baseline vLLM implementation suffers from progressively decaying GPU compute utilization as faster reasoning paths in a batch finish early, \fasttts{} maintains a higher and more consistent utilization by speculatively generating tokens for completed beams. The overall latency is also reduced, as the speculative tokens generated in one iteration can be used as a head start for the next, shortening their required generation time.
The performance of Speculative Beam Extension is also affected by its truncation ratio, \textit{R}. As shown in \figref{fig:spec_beam_ablation} (right), a higher ratio (R=0.85), which aggressively retains speculative work, yields more goodput improvement.

\subsubsection{Effectiveness of Dynamic Prefix-Aware Scheduling}
We evaluate the memory efficiency of Dynamic Prefix-Aware Scheduling against Random and Worst-Case baselines in  \figref{fig:exp_prefix_memory}~(left).
First, KV cache size grows much more slowly with batch size under Dynamic Prefix-Aware Scheduling, indicating higher cache reuse and fewer evictions. The cache size might saturate early due to beam duplication during branching in beam search.
Second, given a fixed KV cache budget, Dynamic Prefix-Aware Scheduling supports substantially larger batches, directly improving throughput.

\subsubsection{Impact of Memory Constraints on Optimizations}
\figref{fig:exp_prefix_memory}~(right) illustrates the performance gains from our optimizations under varying memory availability. The effectiveness of both \textbf{Dynamic Prefix-Aware Scheduling (P)} and its combination with \textbf{Asymmetric Multi-Model Memory Allocation (M+P)} is most pronounced in memory-constrained scenarios. At 1.5 GB of available KV cache, the optimizations deliver substantial goodput gains of 58\% and 145\%, respectively. However, when memory is relatively abundant (e.g., 14 GB), the benefits diminish. This is because a large memory budget can accommodate the entire batch of reasoning paths, which minimizes the KV cache eviction that Dynamic Prefix-Aware Scheduling is designed to prevent. Similarly, when memory is not a bottleneck, a sophisticated allocation strategy becomes less critical.



\section{Related Work}\label{sec:relatedworks}

\subsection{Reasoning Systems and Speculative Execution}

\textbf{Edge and Memory-Constrained Serving:} The recent development of edge LLM serving systems focuses mainly on optimizing the deployment of \textbf{non-reasoning workloads}~\cite{zheng2025review, song2024powerinfer, yu2024edge, wei2025agent}.

\textbf{Efficient Reasoning:}
Although Certaindex~\cite{fu2024efficiently} covers LLM reasoning serving, it focuses solely on CoT reasoning with query-level scheduling and early termination. It does not handle the irregular computation patterns within TTS reasoning trajectories, nor does it optimize scheduling between the generator and verifier.
Beyond system-level serving optimizations, recent algorithmic advances such as speculative reasoning~\cite{pan2025specreason, fu2025scaling, liao2025reward} aim to accelerate inference by leveraging an efficient draft model. 
However, these approaches modify the output distribution and therefore lack algorithmic equivalence. 
We emphasize that such algorithmic techniques are orthogonal to our system-level optimizations and can be seamlessly integrated into our framework to achieve further speedups.

\textbf{Speculative Execution:} The philosophy of speculative generation for LLMs was initially explored at the algorithmic level, primarily through speculative and parallel decoding techniques~\cite{hu2025speculative, li2024eagle, li2025eagle, qin2025dynamic, chengspecinfer-old, fu2023lookahead, cai2024medusa, chen2024hardware}. However, these methods serve as algorithmic enhancements aimed at accelerating LLM decoding by enabling multi-token generation.
Speculative generation has also been applied in retrieval-augmented generation (RAG) systems~\cite{jin2024ragcache, zhang2024accelerating, hu2025hedrarag}, where it is used to prefetch or cache retrieved documents to reduce latency.
In contrast, our Speculative Beam Extension targets the unique system-level bottleneck of parallel reasoning: hardware underutilization caused by irregular step lengths. Instead of verifying draft tokens, we utilize idle compute slots from completed paths to speculatively extend future reasoning steps, thereby mitigating straggler effects and maintaining high GPU occupancy without altering the algorithm. 

\subsection{Memory Management and Prefix Sharing}
Beyond the various memory and scheduling optimizations~\cite{kwon2023efficient, zhong2024distserve, wu2024loongserve, sun2024llumnix, lee2024infinigen, fu2024serverlessllm, lin2024parrot, kim2025oaken}, two other methods have emerged for improving memory efficiency in LLM serving: offloading and prefix sharing.

\textbf{Offloading for LLMs:} 
Offloading is a primary strategy for alleviating LLM memory pressure, typically via computation offloading and data offloading.
In computation offloading, several approaches, such as FlexGen~\cite{sheng2023flexgen}, FastDecode~\cite{he2024fastdecode}, PowerInfer~\cite{song2024powerinfer}, and NEO~\cite{jiang2024neo}, distribute LLM pipelines across CPU and GPU to reduce GPU load. LIA~\cite{kim2025lia} explores offloading to Intel AMX, while~\cite{yuan2025local} supports edge--cloud partitioning for inference.
In data offloading, DeepSpeed-Inference~\cite{aminabadi2022deepspeed} utilizes CPU host memory for activation offloading, while \cite{alizadeh2024llm} explores flash-based offloading strategies. 
AIF~\cite{lee2025aif} pushes this further by supporting in-flash processing to reduce data movement overhead. 



\textbf{Prefix Caching and Sharing:} Most prior work on prefix caching and sharing focuses on query-level optimization using tree-structured management~\cite{kwon2023efficient, zheng2024sglang}.
BatchLLM~\cite{zheng2024batchllm} explores the global prefix reuse through ahead-of-time prefix identification.
FastTree~\cite{panfasttree} improves tree-structured inference via context-query grouping to enhance cache locality.
RAGCache~\cite{jin2024ragcache} investigates prefix sharing with dynamic overlapping between the retrieval and inference steps for retrieval-augmented generation.
KVFlow~\cite{pan2025kvflow} further advances prefix scheduling in multi-agent systems by introducing workflow-aware eviction policies and overlapped KV prefetching.
However, previous work on prefix optimization has primarily focused on \textbf{coarse-grained, query-level sharing} in non-reasoning LLM serving scenarios.
In contrast, LLM reasoning workloads introduce new opportunities for fine-grained,  prefix-aware sharing during decoding.

\section{Conclusion}
This paper presents \fasttts{}, a plug-and-play third-party serving system that makes Test-Time Scaling both practical and efficient on memory-constrained edge devices. 
By analyzing the common execution pattern of mainstream TTS methods, we identify several key system challenges: \textit{i)} hardware underutilization of irregular reasoning search paths, \textit{ii)} suboptimal cache reuse, and \textit{iii)} memory pressure from multi-model execution. 
\fasttts{} addresses these challenges through three novel synergistic techniques: Speculative Beam Extension, Dynamic Prefix-Aware Scheduling, and Asymmetric Multi-Model Memory Allocation. 
Our evaluation shows that \fasttts{} enables low-latency, high-quality reasoning using edge LLMs for memory-constrained devices. It narrows their performance gap with cloud-scale models, and advances the vision of democratized agentic AI.

\section{Acknowledgement}
The support of UKRI and EPSRC (grant numbers UKRI256, EP/V028251/1, EP/N031768/1, EP/S030069/1, and EP/X036006/1), KIAT, AMD and Intel is gratefully acknowledged.


\bibliographystyle{ACM-Reference-Format}
\bibliography{refs}

\appendix
\section{Proof of Optimality for Prefix-Aware Scheduling} \label{app:prefix}

This appendix provides the formal proof that the greedy scheduling algorithm is optimal under a simplified set of assumptions that reflect a highly memory-constrained environment.

\subsection{Assumptions} \label{app:prefix_assumptions}

We make the following list of assumptions to facilitate our formulation and analysis in \secref{sec:prefix}.

\begin{enumerate}
  \item \textbf{Constant Total Work}: For a given set of CoTs to be scheduled in a single TTS iteration, the total number of unique beams (nodes) is fixed. This allows the problem of minimizing eviction cost, $\sum (\text{Nodes}(T_i) - P(T_i, T_{i+1}))$, to be simplified to maximizing the shared prefix sum, $\sum P(T_i, T_{i+1})$.
    \item \textbf{No Preemption During Execution}: The schedule is determined once per TTS iteration, and the execution of the CoTs is non-preemptive.
    \item \textbf{Homogeneous Generation Length}: The number of tokens generated for each beam within a single scheduling cycle is uniform.
\end{enumerate}

\subsection{Proof of Local Optimality}
\label{app:prefix_proof}

\paragraph{Assumptions}
\textbf{Single CoT Batches}: The KV cache has a limited capacity such that only a single CoT can fit into memory at one time. This simplifies the problem by making each Trie, $T_i$, equivalent to a single CoT, $c_i$.

\paragraph{Objective}
We prove that the greedy schedule, $S_G$, is locally optimal. A schedule is defined as locally optimal if its total score cannot be improved by a single swap of any two elements. The surrogate score for a schedule $S = (c_1, c_2, \dots, c_L)$ is the sum of shared prefixes between consecutive elements as mentioned in \secref{sec:prefix}:
\[
\text{Score}(S) = \sum_{k=1}^{L-1} P(c_k, c_{k+1})
\]
We will show that for a schedule $S'$ created by swapping two elements in $S_G$, the change in score, $\Delta \text{Score} = \text{Score}(S') - \text{Score}(S_G)$, is non-positive ($\le 0$).

\paragraph{The Greedy Invariant}
The proof is based on the greedy invariant in \secref{sec:prefix}. 
Formally:
\[
P(c_{k-1}, c_k) = \max_{c_m \in Q} P(c_{k-1}, c_m)
\]

\paragraph{The Interchange Argument}
Consider the greedy schedule $S_G$ and a new schedule $S'$ created by swapping two elements, $c_i$ and $c_j$, where $i < j-1$. Our goal is to show that $S'$ is no better than $S_G$.

\begin{itemize}
    \item \textbf{Greedy Schedule ($S_G$):} \newline 
    $S_G = (\dots, c_{i-1}, \mathbf{c_i}, c_{i+1}, \dots, c_{j-1}, \mathbf{c_j}, c_{j+1}, \dots)$
    \item \textbf{Swapped Schedule ($S'$):} \newline 
    $S' = (\dots, c_{i-1}, \mathbf{c_j}, c_{i+1}, \dots, c_{j-1}, \mathbf{c_i}, c_{j+1}, \dots)$
\end{itemize}

The change in the total score is derived from the four connections affected by the swap.
\[
\Delta \text{Score} = 
\underbrace{\big[ P(c_{i-1}, c_j) - P(c_{i-1}, c_i) \big]}_{\text{Term A}}
\]
\[
+ \underbrace{\big[ P(c_j, c_{i+1}) - P(c_i, c_{i+1}) \big]}_{\text{Term B}}
\]
\[
+ \underbrace{\big[ P(c_{j-1}, c_i) - P(c_{j-1}, c_j) \big]}_{\text{Term C}} 
\]
\[
+ \underbrace{\big[ P(c_i, c_{j+1}) - P(c_j, c_{j+1}) \big]}_{\text{Term D}}
\]

To show that $\Delta \text{Score} \le 0$, we demonstrate that each term in the expression is non-positive. We provide the argument for Term A; a symmetric argument holds for the remaining terms.

By the greedy invariant, 
\[
P(c_{i-1}, c_i) = \max_{c_m \in Q} P(c_{i-1}, c_m) \ge 
P(c_{i-1}, c_j)
\]
Hence, 
\[
P(c_{i-1}, c_j) - P(c_{i-1}, c_i) \le 0
\]

Since all four terms are non-positive, their sum must also be non-positive. Therefore, $\Delta \text{Score} \le 0$, and no single swap can improve the score.

\section{Artifact Appendix}

\subsection{Abstract}

FastTTS is a framework for fast test-time scaling on edge devices, leveraging speculative beam extension, prefix caching, and adaptive memory allocation. The artifact includes the complete source code, benchmark scripts, and configuration files to reproduce the goodput, latency, and accuracy results presented in the paper. It supports evaluating AIME 2024 and AMC 2023 datasets using automated workflows.

\subsection{Artifact check-list (meta-information)}

{\small

\begin{itemize}

  \item {\bf Algorithm: } FastTTS

  \item {\bf Program: } Python 3.11

  \item {\bf Model: } Qwen2.5-Math-1.5B-Instruct, Qwen2.5-Math-7B-Instruct, Math-Shepherd-Mistral-7B-PRM, Skywork-o1-Open-PRM-Qwen-2.5-1.5B

  \item {\bf Data set: } AIME 2024, AMC 2023 (via HuggingFace)

  \item {\bf Run-time environment: } Linux, Conda, Python 3.11, CUDA 12.9, PyTorch 2.7.0

  \item {\bf Hardware: } NVIDIA GeForce RTX 4090 GPU (24 GB VRAM)

  \item {\bf Execution: } Automated via \texttt{run\_all\_experiments.py}

  \item {\bf Metrics: } Goodput, Latency, Accuracy

  \item {\bf Output: } JSONL logs, PDF plots

  \item {\bf Experiments: } Verifier-guided Beam Search across datasets to reproduce goodput, latency, accuracy results.

  \item {\bf How much disk space required (approximately)?: } $\sim$200GB (including model weights and environment).

  \item {\bf How much time is needed to prepare workflow (approximately)?: } $\sim$15 minutes

  \item {\bf How much time is needed to complete experiments (approximately)?: } $\sim$60 GPU hours (depending on GPU)

  \item {\bf Publicly available?: } No

  \item {\bf Data licenses (if publicly available)?: } MIT/Apache 2.0 (Dataset dependent)

  \item {\bf Workflow automation framework used?: } Custom Python scripts

\end{itemize}

}


\subsection{Description}

\subsubsection{How to access}

The artifact is available
\href{https://zenodo.org/records/17943373}
{\texttt{here}}.

\subsubsection{Hardware dependencies}

The experiments require a Linux machine with at least one NVIDIA GeForce RTX 4090 GPU (24 GB VRAM).

\subsubsection{Software dependencies}

Described in the code.

\subsubsection{Data sets}

The artifact uses the following datasets, automatically downloaded via the Hugging Face Datasets library:
\begin{itemize}
    \item \textbf{AIME 2024}
    \item \textbf{AMC 2023}
\end{itemize}

\subsubsection{Models}

The following models are automatically downloaded from Hugging Face Hub:
\begin{itemize}
    \item Generators:
    \begin{itemize}
        \item \texttt{Qwen/Qwen2.5-Math-1.5B-Instruct}
        \item \texttt{Qwen/Qwen2.5-Math-7B-Instruct}
    \end{itemize}
    \item Verifiers:
    \begin{itemize}
        \item \texttt{peiyi9979/math-shepherd-mistral-7b-prm}
        \item \texttt{Skywork/Skywork-o1-Open-PRM-Qwen-2.5-1.5B}
    \end{itemize}
\end{itemize}


\subsection{Installation}

For details, please read the \textit{README.md} file in the repository.

\begin{enumerate}
    \item Download the repository.
    \item Create and activate the Conda environment:
    \begin{verbatim}
conda env create -f environment.yml
conda activate FastTTS
    \end{verbatim}
    \item Install the package in editable mode:
    \begin{verbatim}
pip install -e .
    \end{verbatim}
    \item Install the modified Skywork inference module:
    \begin{verbatim}
cd modified-skywork-o1-prm-inference
pip install -e .
cd ..
    \end{verbatim}
\end{enumerate}


\subsection{Experiment workflow}

The main results can be reproduced using:
\begin{center}
\texttt{run\_all\_experiments.py}
\end{center}

To run all experiments (AIME and AMC datasets, multiple model sizes, varying $N$):
\begin{verbatim}
python run_all_experiments.py --exp
\end{verbatim}
This command runs the benchmark configurations defined in \texttt{benchmarks/configs/}.

To generate the plots from the collected results:
\begin{verbatim}
python run_all_experiments.py --plot
\end{verbatim}

To run both experiments and plotting in sequence:
\begin{verbatim}
python run_all_experiments.py --exp --plot
\end{verbatim}


\subsection{Evaluation and expected results}

Upon successful execution, the script generates the following PDF figures in \texttt{benchmarks/benchmark\_results/figs/}:
\begin{itemize}
    \item \texttt{main\_results\_combined.pdf}: Compares Goodput across different methods and model sizes. It corresponds to Figure~\ref{fig:exp_main}.
    \item \texttt{latency\_combined.pdf}: Shows the latency breakdown. It corresponds to Figure~\ref{fig:exp_latency}.
    \item \texttt{acc.pdf}: Displays the accuracy metrics for the evaluated benchmarks.
    It corresponds to Figure~\ref{fig:top1_acc}.
\end{itemize}
The results should match the trends reported in Figure 12, Figure 13, and Figure 14a of the paper, demonstrating the efficiency improvements. 


\subsection{Experiment customization}

Experiments can be customized by modifying the YAML configuration files located in \texttt{benchmarks/configs/}.
Individual benchmarks can be run via
\texttt{benchmarks/run\_benchmarks.py}
with a specific config file.


\subsection{Notes}

Ensure that the \texttt{conda} environment is activated before running any scripts. The initial run may take longer due to model downloads.

\end{document}